\documentclass{article}
\usepackage[utf8]{inputenc}
\usepackage{amsmath}
\usepackage{booktabs}
\usepackage{graphicx}
\usepackage{tikz}
\usepackage{cancel}
\usepackage{soul}
\usepackage{subcaption}
\usepackage{pgfplots}
\usepackage{csvsimple}
\usetikzlibrary{arrows}
\usetikzlibrary{plotmarks}
\usetikzlibrary{patterns}
\usetikzlibrary{decorations.markings}
\pgfplotsset{compat=1.16}
\usetikzlibrary{arrows}
\usetikzlibrary{positioning,calc}
\usepackage{multirow}
\usepackage{hyperref}
\usepackage{xcolor}
\usepackage{amsmath}
\usepackage{booktabs}
\usepackage{graphicx}
\usepackage{tikz}
\usepackage{subcaption}
\usepackage{authblk}
\usepackage{pgfplots}
\usepackage{natbib}
\usepackage{csvsimple}
\usetikzlibrary{arrows}
\usetikzlibrary{plotmarks}
\usetikzlibrary{patterns}
\usetikzlibrary{decorations.markings}
\pgfplotsset{compat=1.16}
\usetikzlibrary{arrows}
\usetikzlibrary{positioning,calc}
\usepackage{multirow}
\usepackage{hyperref}

\definecolor{lightblue}{RGB}{145, 215, 227}
\definecolor{lightorange}{RGB}{250, 206, 147}

\usepackage{algorithm,algpseudocode}
\usepackage{url}

\makeatletter
\newcounter{phase}[algorithm]
\newlength{\phaserulewidth}
\newcommand{\setphaserulewidth}{\setlength{\phaserulewidth}}
\newcommand{\phase}[1]{%
  \vspace{-1.25ex}
  \Statex\leavevmode\llap{\rule{\dimexpr\labelwidth+\labelsep}{\phaserulewidth}}\rule{\linewidth}{\phaserulewidth}
  \Statex\strut\refstepcounter{phase}\textit{Phase~\thephase~--~#1}
  \vspace{-1.25ex}\Statex\leavevmode\llap{\rule{\dimexpr\labelwidth+\labelsep}{\phaserulewidth}}\rule{\linewidth}{\phaserulewidth}}
\makeatother

\setphaserulewidth{.7pt}

\newcommand{\TR}{{\cal D}_{tr}}
\newcommand{\TS}{{\cal D}_{ts}}
\newcommand{\T}{{\cal T}}

\title{Supervised Feature Compression based on Counterfactual Analysis}
\date{}
\author[1]{Veronica Piccialli}
\author[2]{Dolores Romero Morales}
\author[3]{Cecilia Salvatore}

\affil[1]{Department of Computer Control and Management Engineering Antonio Ruberti, Sapienza University of Rome, Rome, Italy}
\affil[2]{Department of Economics, Copenhagen Business School, Frederiksberg, Denmark}
\affil[3]{Department of Civil Engineering and Computer Science, University of Rome Tor Vergata, Rome, Italy}

\begin{document}

\maketitle
\noindent \textit{The work presented in this paper is published on European Journal of Operations Research, doi: \url{https://doi.org/10.1016/j.ejor.2023.11.019}.}

\section*{Abstract}
Counterfactual Explanations are becoming a de-facto standard in post-hoc interpretable machine learning. For a given classifier and an instance classified in an undesired class, its counterfactual explanation corresponds to small perturbations of that instance that allows changing the classification outcome. This work aims to leverage Counterfactual Explanations to detect the important decision boundaries of a pre-trained black-box model. This information is used to build a supervised discretization of the features in the dataset with a tunable granularity. Using the discretized dataset, an optimal Decision Tree can be trained that resembles the black-box model, but that is interpretable and compact. Numerical results on real-world datasets show the effectiveness of the approach in terms of accuracy and sparsity. \par\smallskip
\textbf{Keywords:}
Machine Learning; Supervised classification; Interpretability; Feature Compression; Counterfactual Analysis

\section{Introduction}

Classification systems based on Machine Learning algorithms are often used to support decision-making in real-world applications such as healthcare \citep{babic2021beware}, credit approval \citep{SILVAEJOR22,KOZODOI20221083,BASTOS2022386,DUMITRESCU20221178}, or criminal justice \citep{ridgewayNIJJ13}. These systems often act as black-boxes that lack of interpretability. Making Machine Learning systems trustworthy has become imperative, and  interpretability, robustness, and fairness are often essential requirements for deployment \citep{EUwhitepaperAI20,goodman2017european,rudin2022interpretable}.

This paper is devoted to enhancing the interpretability of black-box classifiers. Without loss of generality, we will focus on binary classification problems. We are given a black-box classification model, hereafter, the target model. The goal is to discretize the features by detecting their most critical values.
Discretization techniques have often been proposed in the literature as a preprocessing step that allows the transformation of continuous data into categorical ones \citep{DOUGHERTY1995194, dash2011comparative, 6152258, https://doi.org/10.1002/widm.1173}; the objective is to make the representation of the knowledge more concise. Note that a trivial way to discretize continuous features consists in creating a dummy variable for each split point along a feature; a split point is defined as the middle value between two consecutive points on a feature. However, this results in a very large number of dummy variables. In our approach we aim at a meaningful compressed feature representation, that gives us a more interpretable view of the data since the only relevant values are the critical ones defining the discretization.  
In addition, this procedure acts as a Feature Selection technique \citep{PIRAMUTHU2004483}, meaning that features on which no critical values are extracted are considered not important for detecting the input-output relationship, and can thus be filtered out. Discretizing features can also help in reducing noise that can cause overfitting to  training data. Finally, we make it more affordable to train an optimal classification tree \citep{carrizosaTOP21} as the interpretable surrogate of the target model, and to certify its optimality, thanks to the reduction in the number of thresholds that need to be considered, namely, the critical ones.

In this paper, the feature discretization will be supervised by a Counterfactual Analysis on the target model. Counterfactual Analysis is a post-hoc local explainability technique \citep{karimi2022survey,martensMISQ14,molnar2020interpretable,wachter2017counterfactual} that has gained a lot of attraction especially in Supervised Classification. The starting point is an already trained classification model and an observation that has been classified by the model in the undesired class, e.g., as a \emph{bad} payer in a credit approval application \citep{FETHI2010189,DOUMPOS2022}. Counterfactual Analysis provides feedback on how to change the features of the observation in order to change the prediction given by the model to the desired class, e.g., as a \emph{good} payer in the credit approval application above. The Counterfactual Explanation depends on the cost function used to measure the changes, which is minimized, and the constraints imposed on the explanation, such as upper bounds on the changes to continuous variables or the correct modeling of changes to categorical features.

We propose to use Counterfactual Analysis to detect decision boundaries of the target model. For this, we extract a set of univariate decision boundaries, i.e., axis-parallel hyperplanes, from Counterfactual Explanations. We use observations that have been correctly classified by the target model. To each of those, we find their counterfactual explanation, yielding a set of axis-parallel hyperplanes. By putting together all these hyperplanes from all the counterfactual explanations, we can identify for each feature a set of thresholds. This step allows us to derive a meaningful supervised discretization of the original dataset, where the cutting points on each feature are the thresholds, and hence are related to the decision boundary of the target model. With this, we can reproduce an equivalent decision boundary by means of a classification tree. 

A major strength of our approach is that our supervised discretization is based on an optimization problem, the counterfactual problem. First, this allows us to control the importance of the features extracted, and hence the granularity of the discretization, through the counterfactual cost function. Furthermore, we can impose desirable properties of the boundary, by modelling constraints on the counterfactuals, like plausibility or fairness \citep{carrizosaGCE23,maragno2022counterfactual} 

In summary, given a black-box classification model, Counterfactual Explanations can help us to find a supervised discretization where the compression focuses on the important decision boundaries of the target model. With this new representation of the data we can build an interpretable surrogate model, namely a optimal univariate decision tree. Our approach has the following advantages:

\begin{description}
\item We produce a supervised discretization of the original dataset whose granularity can be tuned. The discretization is driven by an optimization problem, the counterfactual problem, that allows us to control the importance of the thresholds extracted and to impose properties on the boundary. 
\item The interpretable model that we build uses thresholds extracted from the black-box model. This guarantees the use of features and thresholds that are relevant for the problem, since they represent the decision boundaries of a more complex classification model.
\item Our methodology is suitable for any type of classifier for which the counterfactual problem can be solved efficiently, such as tree ensemble models (e.g., random forests, gradient boosting) or linear models (e.g., logistic regression, linear support vector machines).
\item Our approach allows to strongly compress the original dataset, making it feasible to train an optimal classification tree and to certify its optimality \citep{pmlr-v119-lin20g}. The reduced dataset can be used to train other machine learning models.
\item Once all the thresholds have been computed using a  counterfactual analysis, the user can easily visualize the accuracy and granularity tradeoff by changing one parameter of the discretization. 
Tuning the granularity just results in selecting a subset of the thresholds computed by counterfactual analysis and is thus computationally cheap.
\item Our discretization shows a good tradeoff between out-of-sample accuracy and standard metrics for evaluating discretization procedures \citep{6152258}, such as compression rate and inconsistency rate.
\end{description}

The remainder of the paper is structured as follows. In Section \ref{sec:lit_rev} we analyze the existing literature on Decision Trees, Counterfactual Explanations and discretization procedures. In Section \ref{sec:meth} we formalize our method. In Section \ref{sec:exp_setup} we describe the experimental setup and the obtained results. Finally, in Section \ref{sec:concl} we draw some conclusions, and propose some lines for future research.

\section{Literature Review}\label{sec:lit_rev}

The most popular and inherently interpretable models are univariate Decision Trees \citep{carrizosaTOP21}.
Training an optimal Decision Tree is known to be an NP-complete problem \citep{laurent1976constructing}. Many approaches for training Decision Trees thus rely on heuristics; the most used heuristic is based on a top-down greedy strategy in which the tree structure is recursively grown from the root to the terminal nodes \citep{Breiman1984CART, quinlan1986induction, quinlan2014c4}. However, because of their greedy nature, these heuristics may result in poor generalization capabilities.

Recently, there has been an increasing interest in developing mathematical optimization formulations and numerical solution approaches to train Optimal Classification Trees \citep{carrizosaTOP21}. One of the first approaches for Optimal Classification Trees was proposed in \cite{bertsimas2017optimal}, where a Mixed-Integer Linear Programming problem (MILP) for both univariate and multivariate classification trees is formulated. This formulation, however, has a weak linear relaxation and, thus, it is hard to solve to optimality on real-sized instances; the authors thus propose a local search algorithm to try to overcome this issue, returning a local solution that has not guarantee to be the global minimum \citep{dunn2018optimal}. Recently, new approaches based on binary features have been proposed, and ad-hoc algorithms have been designed with the purpose of being scalable and fast \citep{verwer2019learning, pmlr-v119-lin20g, gunluk2021optimal, aghaei2021strong}.
Furthermore, the dynamic programming approach proposed in \cite{pmlr-v119-lin20g} is better suited for realistic datasets with a few continuous features, thanks to a new representation of the dynamic programming search space.

Still, the number of continuous features that can be handled is limited as each continuous feature is associated with a large number of dummy variables, thus enlarging the search space and slowing down the optimization process. In order to extend the applicability of the procedure, the authors propose some strategies to exploit knowledge extracted from black-box models to make the Decision Tree's optimization faster \citep{McTavish_Zhong_Achermann_Karimalis_Chen_Rudin_Seltzer_2022}. The most effective strategy is the ``Guessing Thresholds'', which allows them to limit the number of dummy variables modelling continuous features. The ``Guessing Thresholds'' uses Tree Ensembles as a black-box model, and it considers as thresholds the ones used in the various splits of each tree in the ensemble; the thresholds are sorted by their importance (e.g. their Gini index) so that the least important can be removed; the Tree Ensemble is then re-fitted with the remaining features, and the procedure is repeated until there is a too large drop in the training performance.
The idea of leveraging black-box models, and in particular Tree Ensembles, to build Decision Trees is also proposed in \cite{pmlr-v119-vidal20a}, in which the authors design a dynamic programming approach where a single Decision Tree is built to reproduce exactly the decision function of a Tree Ensemble; however, in general, the resulting Decision Tree can be large and thus can lose some interpretability.

Our research collocates in between the approach in \cite{pmlr-v119-vidal20a} and the one in \cite{McTavish_Zhong_Achermann_Karimalis_Chen_Rudin_Seltzer_2022}. We use optimization as in \cite{pmlr-v119-vidal20a}, but we 
do not reproduce exactly the original black-box model. Instead, we exploit the boundary of the black-box model to compress the original dataset in a meaningful way by means of optimization. Then, we use the compressed dataset to efficiently train an optimal decision tree as in \cite{McTavish_Zhong_Achermann_Karimalis_Chen_Rudin_Seltzer_2022}. We force interpretability by limiting the depth of the optimal decision tree. Our discretization is driven by a procedure based on optimization for computing Counterfactuals Explanations, that are local explainability techniques, that can be used to explain a single decision of a black-box model.
Indeed, Counterfactual Explanations allow providing feedback to users on how to change their features in order to change the outcome of the decision \citep{karimi2022survey, guidotti2022counterfactual}.
Formally, the Counterfactual Explanation of a datapoint $x^0$, namely $x^{CE}$, is defined as the perturbation of minimal cost (w.r.t. some cost function) that allows changing the classification outcome. An optimization problem for computing the Counterfactual Explanations $x^{CE}$ of a given point $x^0$, namely the counterfactual problem was proposed in \cite{wachter2017counterfactual}:
\begin{align}
    \text{arg}&\min_{x^{CE}}C(x^0,x^{CE})\quad\text{s.t. } f(x^{CE})=y^{CE},\label{eq:CE}
\end{align}
where $C$ is a cost function, $f$ is the classification function and $y^{CE}\neq f(x^0)$ is the required label for the Counterfactual Explanation. The problem is then reformulated as an unconstrained problem with a differentiable objective function, composed of two terms:
the first term of the objective requires the classification function to be as close as possible to the required label $y^{CE}$, while the second term requires minimizing the distance between the Counterfactual Explanation and the initial point.

Later in the literature, \cite{verma2022counterfactual} identifies some additional constraints Counterfactual Explanations should satisfy.
Proximity requires that a valid Counterfactual Explanation must be a \textit{small} change with respect to the initial point. Actionability implies that the Counterfactual Explanation can modify some features (e.g., income), while others must be \textit{immutable} (e.g., sex, race).
Sparsity requires a Counterfactual Explanation to be \textit{sparse}, i.e. as few features as possible should change. This makes the Counterfactual Explanation more effective because simpler explanations can be better understood by users. Data Manifold Closeness suggests that a Counterfactual Explanation should be \textit{realistic}, meaning that it should be close to training data. Finally, Causality requires that the Counterfactual Explanation adheres to observed \textit{correlations} between features. Examples of constraints modelling domain knowledge and actionability can be found in \cite{pmlr-v139-parmentier21a,maragno2022counterfactual}. Robustness is also an important requirement, as it imposes that Counterfactual Explanations remain valid after retraining the model or slightly perturbing the input features
 \citep{forel2022robust,fernandez2022factual,maragno2023finding}.

For some classifiers such as linear Support Vector Machines (SVMs) or Tree Ensembles, it is possible to derive an explicit expression of the classification function $f$, allowing to directly write problem \eqref{eq:CE} as a Linear Programming problem or at most a convex quadratic problem. Integer variables can be added to represent the $l_0$-norm in the objective function. For SVMs with non-linear kernels or for Neural Networks it is not possible to do so. To overcome this issue, \cite{maragno2022counterfactual} suggest the idea that the Counterfactual Explanation problem is a special case of Optimization with Constraint Learning, in which some of the constraints are learnt through a predictive model.

Recent literature states that Counterfactual Explanations can provide useful insights into the classification model, allowing them to be used not only for post-hoc explainability but also for debugging and detecting bias in models \citep{sokol2019counterfactual}. For example, if it turns out that without imposing the actionability constraints the Counterfactual Explanation changes a sensitive feature (e.g., gender) by saying that a woman would receive the loan if she were a man, then the classification model may be biased. This observation highlights that Counterfactual Explanations can be used to detect biases in Machine Learning models, opening the possibility of designing new fairness metrics that rely on Counterfactual Explanations \citep{kusner2017counterfactual,goethals2023precof}.

Some recent literature \citep{9555622,mothilal2020explaining,aivodji2020model,zhao2021exploiting} focuses on the uses of Counterfactual Explanations in an adversarial setting for detecting the decision boundaries of machine learning models.

In this work, we propose to use Counterfactual Explanations to enhance the interpretability
of the model itself.
We show that using Counterfactual Explanations to detect the important decision boundaries of a black-box model allows us to design a supervised discretization technique that helps to build a optimal Decision Tree. 
Using a discretization technique as a preprocessing step can lead to many advantages: (1) some Machine Learning algorithms prefer categorical variables, e.g., the Naive Bayes classifier \citep{yang2009discretization,flores2011handling}; (2) discretized data are easier to understand and to explain; (3) discretization can decrease the granularity of data, potentially decreasing the noise in the dataset  \citep{6152258}.
Nevertheless, any discretization process generally leads to a loss of information, making it crucial the minimization of the loss of information.
In \cite{6152258}, a taxonomy for categorizing discretizing methods is introduced. The effectiveness of a discretization procedure can then be evaluated according to different aspects: (a) the discretization should be able to compress the information as much as possible, by detecting as few intervals as possible; (b) the inconsistency rate produced by the discretization, i.e. the unavoidable error due to multiple points associated with the same discretization but with different labels; (c) the classification rate obtained by an algorithm trained on discretized data compared with one provided by the same algorithm on the initial representation of the dataset.

\section{Method}\label{sec:meth}
We assume we have at hand a binary-classification training set: \[\TR=\{(x_i,y_i):x_i\in \Re^m, y_i\in\{0,1\} \quad\forall i=1\dots n_{tr}\},\] taken out from a sample of $n$ individuals. This procedure can, however, be easily extended to multi-class problems. For simplicity, we assume w.l.o.g. all features to be scaled between 0 and 1.

We train a black-box model $\T$ using $\TR$, that we use as the target model for our procedure. For our purpose, $\T$ can implement any classification algorithm as long as we can efficiently compute the Counterfactual Explanation associated with each instance $x\in\TR$ with respect to model $\T$. Our objective is to train a small, compact and interpretable decision tree that acts as a surrogate for $\T$. At each branch node $s$, a univariate Decision Tree  takes a univariate decision: if the input $x$ on a given feature $j$ is less or equal than a threshold $\tau_s$ the point is directed to the left child of $s$, otherwise to the right child. By following the path of each point $x$ from the root of the tree to the last level, a set of points is assigned to each leaf $l$ of the tree; the classification outcome at each leaf $l$ depends on the most frequent label of the points assigned to $l$. Training a Decision Tree results in choosing which feature to consider at each node $s$ and the value of the threshold $\tau_s$.

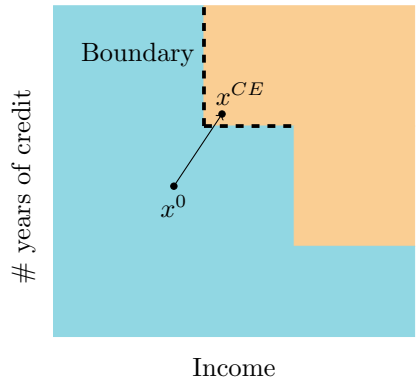
\begin{figure}
    \centering
    \begin{tikzpicture}[scale=0.8]
        \filldraw[color = lightblue, fill = lightblue] (1,1) rectangle (3.5,-1);
        \filldraw[color = lightorange, fill = lightorange] (3.5,1) rectangle (7,-1);
        \filldraw[color = lightblue, fill = lightblue] (1,-1) rectangle (5,-3);
        \draw[white] (0.5,0)--(0.5,-2) node[rotate = 90]{\textcolor{black}{\# years of credit}};
        \draw[white] (2.5,-5)--(4,-5) node {\textcolor{black}{Income}};
        \filldraw[color = lightorange, fill = lightorange] (5,-1) rectangle (7,-3);
        \filldraw[color = lightblue, fill = lightblue] (1,-3) rectangle (7,-4.5);
        \filldraw [black] (3,-2) circle (1.5pt) node[anchor = north]{$x^0$};
        \draw[->](3,-2) -- (3.8,-0.8);
        \filldraw [black]  (3.8,-0.8) circle (1.5pt) node[anchor = south, xshift=7pt]{$x^{CE}$};
        \draw[line width=0.5mm,dashed] (3.5,-1)--(3.5,1) node[anchor=north,yshift=-10pt, xshift=-25pt]{Boundary};
        \draw[line width=0.5mm,dashed] (3.5,-1)--(5,-1) node[anchor=east,yshift=7pt, xshift=-12pt]{};
    \end{tikzpicture}
    \caption{Closeness of a Counterfactual Explanation to the Decision Boundaries of the Target model produced by a Random Forest projected on two features (Income and \# years of credit).}
    \label{fig:boundaries}
\end{figure}

\begin{figure}
\centering
\begin{subfigure}{.45\textwidth}
  \centering
  \includegraphics[width=\linewidth]{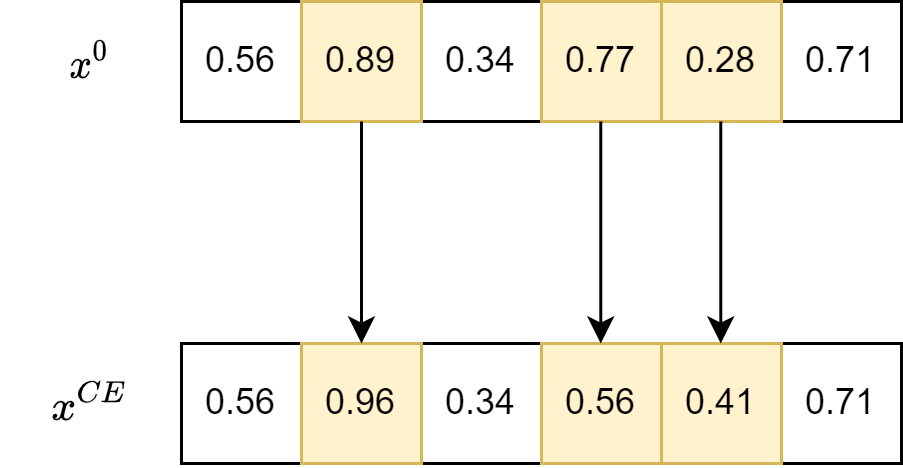}
  \caption{An initial point $x^0$ and its counterfactual explanation}\label{fig:thresholds_example}
\end{subfigure}\hspace{10pt}
\begin{subfigure}{.45\textwidth}
  \centering
    \includegraphics[width=.9\linewidth]{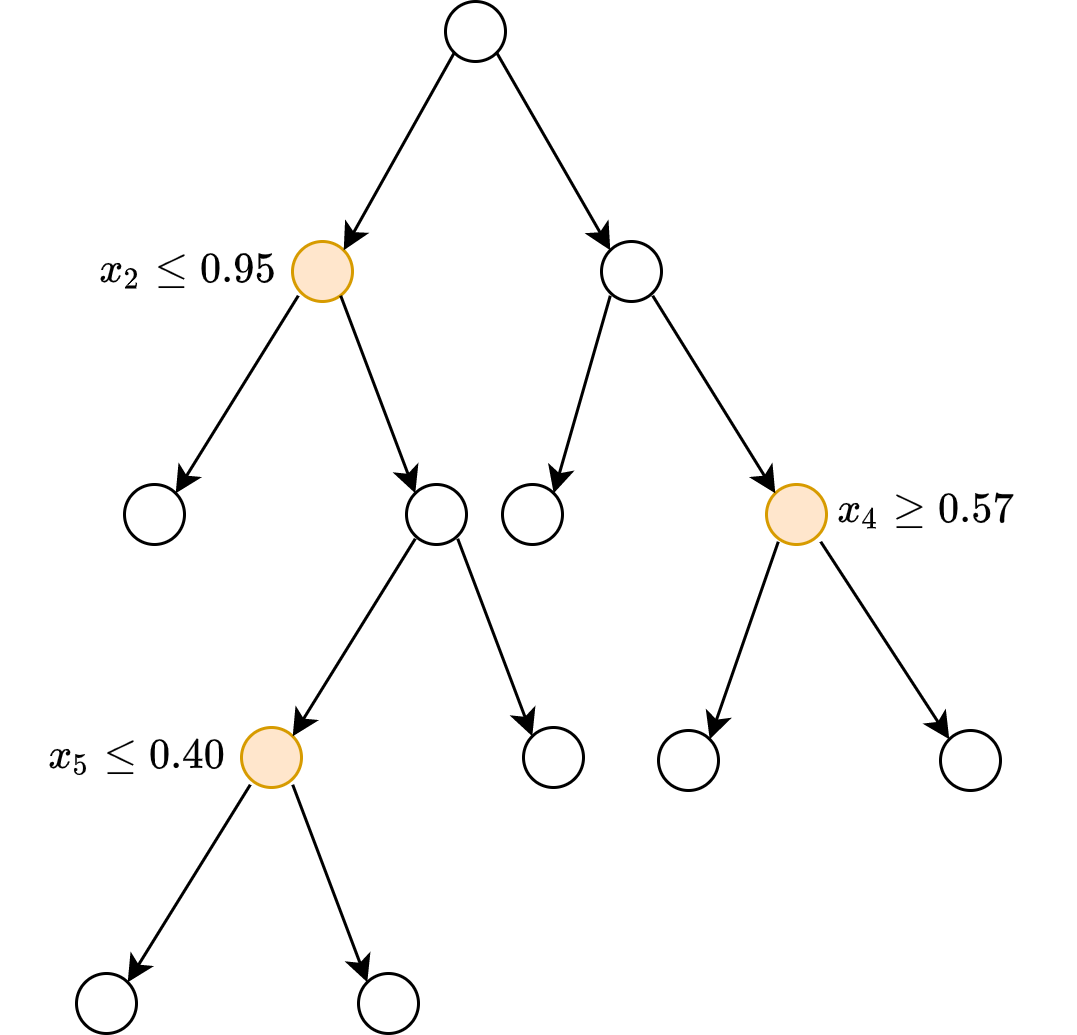}
  \caption{An example of a decision tree}
  \label{fig:tree_ex}
\end{subfigure}
\caption{Assume that $x^{CE}$ is computed by solving problem \eqref{eq:CE} with $x^0$ as input for a given Tree Ensemble acting as a target model. A perturbation ($\pm \epsilon$) of the values of the features where $x^0$  and $x^{CE}$ differ can be used as splitting values for the nodes of a univariate Decision Tree.}
\end{figure}

As shown in Figure \ref{fig:boundaries}, the intuition behind our procedure is that a Counterfactual Explanation is very close ($\pm\epsilon$) to some decision boundaries of the target model. So, if we generate a large set of Counterfactual Explanations, we should be able to identify the most critical decision boundaries of the model and use them to guide the training procedure of the surrogate Decision Tree. Each Counterfactual Explanation $x^{CE}$ perturbs just a subset of the features of the corresponding initial point $x^0$ (Figure \ref{fig:thresholds_example}); the values these features assume can be used to mimic some decision boundaries of the Target model, i.e. these values can be used as splitting values in the nodes of the surrogate Decision Tree (Figure \ref{fig:tree_ex}).

Following \cite{carrizosacounterfactual21} for computing Counterfactual Explanations, we solve a problem derived from \eqref{eq:CE} under the assumption that the Target model is a Tree Ensemble. This problem takes the following parameters: $y^{CE}\in \{0,1\}$, with $y^{CE}\neq y^0$,  is the required outcome for the Counterfactual Explanation; $T$ is the set of trees in the Tree Ensemble; ${\cal L}_t$ and ${\cal N}_t$ denote respectively the leaves and the internal nodes of each tree $t$ in the tree ensemble; $v_{t,s}$ and $c_{t,s}$ denote respectively which feature and which threshold is used to split at node $s$ in tree $t$; ${\cal A}_L(t,l)$ and ${\cal A}_R(t,l)$ denote respectively the ancestors $s$ of leaf $l$ in tree $t$ whose left/right path leads from $s$ to $l$. We denote by $p_k(x)$ the probability for point $x$ of being in class $k$; the specific expression of $p_k(x)$ depends on the tree ensemble method used (e.g. Random Forest \citep{breiman2001random} or Gradient Boosting \citep{chen2015xgboost}). The variables define the point $x^{CE}$ and a binary variable $z_{t,l}$ that denotes the assignment of $x^{CE}$ to one of the leaves for each of the trees in the tree ensemble. The formulation is the following:
\begin{align}
    \min_{x^{CE}}\ & C(x^0, x^{CE}) := \lambda_0\|x^0-x^{CE}\|_0 + \lambda_1\|x^0-x^{CE}\|_1 + \lambda_2\|x^0-x^{CE}\|^2_2\label{eq:CE_obj}\\
    &x^{CE}_{v_{t,s}} - M_0(1-z_{t,l})+\epsilon_{v_{t,s}}\leq c_{t,s}\qquad\forall t\in T,l\in{\cal L}_{t},s\in{\cal N}_{t}\colon s\in{\cal A}_{L}(t,l)\label{eq:CE_class_assignment_a}\\
    &x^{CE}_{v_{t,s}} + M_1(1-z_{t,l})-\epsilon_{v_{t,s}}\geq c_{t,s}\qquad\forall t\in T,l\in{\cal L}_{t},s\in{\cal N}_{t}\colon s\in{\cal A}_{R}(t,l)\label{eq:CE_class_assignment_b}\\
    &\sum_{l\in{\cal L}_t}z_{t,l}=1\qquad\qquad\forall t\in T\label{eq:CE_class_assignment_c}\\
    & p_{y^{CE}}(x^{CE})\geq  p_{y^{0}}(x^{CE})+\epsilon\label{eq:CE_class_assignment_d}\\
    & x^{CE}\in{\cal X}^0.\label{eq:CE_actionability}
\end{align}
In the objective function, we consider the weighted combination (with non-negative coefficients) of $l_0$-,$l_1$- and $l_2$-norm; the $l_0$-norm is used for feature sparsity while the $l_1$- and $l_2$- norms are used to measure proximity. The $l_0$- and $l_1$-norms are modelled in a standard way by introducing respectively binary variables and linear constraints.
By increasing the value of parameter $\lambda_0$, we encourage changing as few features as possible. Both $M_0$ and $M_1$ can be set to 1 since the features are assumed to be scaled between 0 and 1.
Constraint \eqref{eq:CE_actionability} imposes that the Counterfactual Explanation should belong to a set ${\cal X}^0$ that represents a plausibility set for the initial point $x^0$; actionability, data manifold closeness, causality and other requirements can thus be expressed via additional constraints.
Equations \eqref{eq:CE_class_assignment_a}-\eqref{eq:CE_class_assignment_d} impose that the label assigned to the Counterfactual Explanation should be the required label $y^{CE}$; in this set of constraints, binary variables $z$ model the assignment of $x^{CE}$ to one of the leaves, for each tree in the tree ensemble. We can notice that constraints \eqref{eq:CE_class_assignment_a} and \eqref{eq:CE_class_assignment_b} depend on a threshold $\epsilon_j$ for each feature $j\in[1\dots m]$. In our paper, we set the value of $\epsilon_j$ as the smallest difference between the two consecutive values that feature $j$ assumes on the datapoints of $\TR$. For the value of $\epsilon$ that appears in constraint \eqref{eq:CE_class_assignment_d}, we use a fixed value, while in \cite{forel2022robust} it is analyzed how to set it for requiring Counterfactual Explanation robustness in Tree Ensembles. The expression of constraint \eqref{eq:CE_class_assignment_d} depends on the specific tree ensemble method. We report the expressions for Random Forest:
\begin{equation}
        \frac{1}{|T|}\sum_{t\in T}\sum_{l\in {\cal L}_t}w_{t,l,y^{CE}}z_{t,l}\geq \frac{1}{|T|}\sum_{t\in T}\sum_{l\in {\cal L}_t}w_{t,l,y^0}z_{t,l}+\epsilon\label{eq:class_assignment_rf},
\end{equation}
where $w_{t,l,k}$ denotes the classification weight of leaf $l$ in tree $t$ for class $k$,
and Gradient Boosting:
\begin{equation}
\begin{cases}
    p_0+lr\sum_{t\in T}\sum_{l\in {\cal L}_t}w_{t,l}z_{t,l}\geq \epsilon\quad&\text{if }y^{CE}=1\\
    p_0+lr\sum_{t\in T}\sum_{l\in {\cal L}_t}w_{t,l}z_{t,l}\leq -\epsilon\quad&\text{if }y^{CE}=0,\\
\end{cases}\label{eq:class_assignment_gb}
\end{equation}
where $p_0$ is the initial prediction of Gradient Boosting computed as the fraction of points in the majority class in the training set, $lr$ is the learning rate, and $w_{t,l}$ is the value predicted by leaf $l$ of tree $t$.

For each counterfactual couple $(x^0, x^{CE})$, composed by a point $x^0$ and its counterfactual explanation $x^{CE}$, we restrict our attention to the features that change significantly, i.e. $|x_j^0-x_j^{CE}|>\epsilon_j$. Then, we can compute a possible splitting threshold for feature $j$:
\begin{equation}
t_j = x^{CE}_j +\epsilon_j*{\mathrm{ sign}}(x^{0}_{j}-x^{CE}_j)\qquad \text{if }|x^{0}_{j}-x^{CE}_{j}|>\epsilon_j.\label{eq:thresholds}
\end{equation}
\noindent Generating a set of counterfactual couples thus results in computing for each feature $j$ a set of thresholds:
\begin{equation}
\tau_j=\{t_j = x^{CE}_j +\epsilon_j*{\mathrm{ sign}}(x^{0}_{j}-x^{CE}_j),\quad \forall (x^{0},x^{CE})\colon |x^{0}_{j}-x^{CE}_{j}|>\epsilon_j\}.
\end{equation}

\noindent The set of thresholds across all features is denoted by ${\tau= \cup_{j\in [1\dots m]}\tau_j}$. Our objective is to identify the most important decision boundaries of $\T$.
The importance of a feature $j$ can be measured by looking at the number of counterfactual couples for which a threshold of feature $j$ is extracted, that is the size of $\tau_j$. Note that increasing parameter $\lambda_0$ in the cost function \eqref{eq:CE_obj} results in principle in selecting the most important features.
Let us denote by $\pi_{t_j}$ the multiplicity of ${t_j\in\tau_j\ \forall j\in [1\dots m]}$. We denote by ${\tau^Q_j=\{t\in\tau_j\colon \pi_t\geq F_Q\}}$ and by ${\tau^Q=\cup_{j\in  [1\dots m]}\tau_j^Q}$, where $Q$ is a quantile value and $F_Q$ is the $Q$-th quantile of the frequency distribution $\pi = {\{\pi_{t_j}}\}_{\forall t_j\in\tau_j,\ \forall j\in [1\dots m]}$.

After fixing a quantile value $Q$, we want to use the thresholds in $\tau^Q$ as splitting values in the nodes of the surrogate Decision Tree. This can be translated into a Feature Discretization procedure of the data in $\TR$. Parameter $Q$ allows thus to control the number of thresholds considered in $\tau^Q$ across all features.

Each feature $j$ can be replaced with a set of binary variables, one for each threshold $t\in\tau_j^Q$. The binary variable $b_{it}$ associated with each threshold $t\in\tau_j^Q$ represents whether feature $j$ for data point $x_{i}$ lies before or after $t$:
\begin{equation}
b_{it} = \begin{cases}
        0\quad&\text{if }x_{ij}\leq t\\
        1\quad&\text{otherwise}
      \end{cases}\qquad\forall t\in\tau_j^Q.
\end{equation}

With this procedure we transform all the numerical features into binary variables. If we have no threshold on a feature, we remove it from the discretized dataset. Note that the higher the $Q$, the more compressed is the discretized dataset.

\subsection{Algorithm}
In this section, we describe the details of our procedure, namely FCCA (Feature Compression based on Counterfactual Analysis). As Target system $\T$ we can consider any black-box model for which it is possible to solve the counterfactual problem \eqref{eq:CE}.

After training the Target system, the second step is to extract from $\TR$ a large set of points $\cal M$ for computing their Counterfactual Explanation. Our numerical experiments showed that the time for solving problem \eqref{eq:CE_obj}-\eqref{eq:CE_actionability} strongly depends on the closeness of the initial point to the decision boundary of $\T$. In fact, if the initial point is far from the decision boundary of $\T$, a large perturbation may be needed to cross the decision boundary i.e. change the classification outcome. A proxy for the time needed for solving problem \eqref{eq:CE_obj}-\eqref{eq:CE_actionability} is thus the classification probability. On the other hand, it is possible that considering points too close to the decision boundary (i.e. classified with a very low probability) can introduce some noise and make the procedure less robust. We can thus define $\cal M$ to be composed by the points in the training set $\TR$ which are correctly classified and where the classification probability is bounded between two values $0.5\leq p_0\leq 1$ and $p_0\leq p_1\leq 1$:
\begin{equation}{\cal M} = \{(x_i,y_i)\in \TR \text{ if }f_{\T}(x_i)=y_i\ \&\ p_0\leq \Pi_{\T}(x_i)\leq p_1 \},\end{equation}
where $f_{\T}(x_i)$ and $\Pi_{\T}(x_i)$ return respectively the classification label and the classification probability for $x_i$. We compute the set $\cal C$ of Counterfactual Explanations for all points in $\cal M$.

We can use equation \eqref{eq:thresholds} for extracting, from all couples in $({\cal M}[i], {\cal C}[i])_{i\in [1\dots |{\cal M}|}]$, the set of thresholds $\tau$. We set a value of $Q$ between 0 and 1 and use this value to restrict $\tau$ to include only the more frequent thresholds $\tau^Q$. The output of the procedure is thus the discretization of both data in the training and test set, namely $\TR'(\tau^Q)$ and $\TS'(\tau^Q)$. The discretized data can later be used to train and test the performance of a Decision Tree, which can be trained both by using the heuristic CART algorithm or by using an optimal approach as the one proposed in \cite{pmlr-v119-lin20g}. Note that training a Decision Tree on the discretized dataset implies that the decision tree uses exactly some of the thresholds found by the counterfactual computation.

The described procedure is summarized in Algorithm \ref{alg:code}.

\subsection{Discretization effectiveness}
As a side product, our procedure returns different discretizations by changing $Q$. In order to evaluate the effectiveness of these discretizations in terms of compression ability, we introduce two metrics:
\begin{description}
    \item[Compression rate] When we apply the discretization, some points collapse to the same discretization. The compression rate is defined as $\eta = 1-r$, where $r$ is the ratio between the number of points in $\TR$ with different discretizations and the total number of points in $\TR$.
    \item[Inconsistency rate] As a downside of the compression, when multiple points collapse to the same discretization, it may happen that not all of them have the same label. For each feature $j$, we denote by $\xi_j$ the number of thresholds in $\tau^{Q}_{j}$; the number of possible values that each discretized point assumes on feature $j$ is thus $\xi_j+1$. The number of possible discretized points ${\cal N}_{\tau^Q}$ thus depends on how many thresholds we have for each feature:
    \begin{equation}{\cal N}_{\tau^Q} = \prod_{j=1}^{m}(\xi_j+1).\end{equation}
    For each possible discretization $l\in [1\dots {\cal N}_{\tau^Q}]$, we denote by $\Omega_l\subseteq\TR$ the set of points that fall into that discretization. We denote by $\Omega_l^0=\{x_i\in\Omega_l\colon y_i=0\}$ and by $\Omega_l^1=\{x_i\in \Omega_l\colon y_i=1\}$. The number of inconsistencies $\delta_l$ in $\Omega_l$ is thus equal to the number of points in $\Omega_l$ with minority label: $\delta_l=\min\{|\Omega_l^0|, |\Omega_l^1|\}$. The inconsistency rate produced by the discretization procedure is thus expressed as: 
    \begin{equation}\delta =\frac{1}{|\TR|}\sum_{l=1}^{{\cal N}_{\tau^Q}}\delta_{\Omega_l}.\end{equation}
    The inconsistency rate represents an irreducible error; thus, $1-\delta$ is an upper bound to the accuracy that any classifier built on the discretized dataset is able to achieve. Note that in principle it is always possible to train on the discretized dataset an univariate decision tree, without limiting the depth, achieving an accuracy on the training set equal to $1-\delta$. 
\end{description}

\noindent Both the compression rate and the inconsistency rate depend on the value of $Q$. For high values of $Q$ we consider a low granularity discretization that results in a high compression rate, related to high interpretability; but on the other hand, the discretization could produce a large number of inconsistencies, that represents a lower bound on the error rate of any classification method on the discretized dataset. The objective is thus to choose $Q$ in order to keep a good trade-off between the compression rate and the inconsistency rate.  Summarizing, in our methodology we have two parameters allowing to control the quality of the discretized dataset: the parameter $\lambda_0$ in the cost function \eqref{eq:CE_obj} influences the number of features involved in the discretization (feature sparsity), whereas the parameter $Q$ controls the number of thresholds involved across all the features (threshold sparsity).

\begin{algorithm}
\begin{algorithmic}[1]

\State \textbf{Input data: } $p_0 \geq 0.5$, \qquad $p_0\leq p_1\leq 1, \qquad \lambda_0,\lambda_1, \lambda_2\geq 0
, \qquad 0\leq Q\leq 1$
\State \textbf{Default: }$p_0:=0.5,\quad p_1:=1,\quad \lambda_0:=0.1,\quad \lambda_1:=1,\quad \lambda_2:=0,\quad Q:=0$
\State $\TR\gets\{(x_i,y_i):x_i\in \Re^m, y_i\in\{0,1\} \quad\forall i=1\dots n_{tr}\}$
\State $\TS\gets\{(x_\ell,y_\ell):x_\ell\in \Re^m, y_\ell\in\{0,1\} \quad\forall \ell=1\dots n_{ts}\}$
\State $\T \gets$ black-box model \texttt{trained on $\TR$}
\State $f_\T \gets$ \texttt{predictions function of $\T$}
\State $\Pi_\T \gets$\texttt{classification probability function of $\T$}

\phase{Computing $\cal M$ and $\cal C$}

\State ${\cal M}=\{(x_i,y_i)\in\TR\colon f_\T(x_i)=y_i\ \&\ p_0\leq \Pi_\T(x_i)\leq p_1\}$
\State ${\cal C}\gets$ \texttt{Counterfactual Explanations of }${\cal M}$ \texttt{with parameters} $\lambda_0, \lambda_1, \lambda_2$

\phase{Computing the thresholds $\tau$}
\State $\tau_j=\{\}\quad\forall j=1\dots m$
\For{$i= 1\dots |\cal M|$}
\State $(x^{0},y^{0})={\cal M}[i]$
\State $(x^{CE},y^{CE})={\cal C}[i]$
\For{$j= 1\dots m$}
\If{$|x^{0}_{j}-x^{CE}_{j}|>\epsilon_j$}
\State $t_j = $ $x^{CE}_j +\epsilon_j*{\mathrm{ sign}}(x^{0}_{j}-x^{CE}_j)$
\State \texttt{add }$t_j$\texttt{ to }$\tau_j$
\EndIf
\EndFor
\EndFor
\State $\pi \gets$ \texttt{frequency distribution of the thresholds in }$\cup_{j=1\dots m}\tau_j$ 

\phase{Discretizing the dataset}
\State $F_Q =$ quantile($\pi$, $Q$)
\State $\tau^Q_j=\{t\in\tau_j\colon \pi_{t}\geq F_Q\}\quad \forall j=1\dots m$
\State $\TR'(\tau^Q)\gets \emptyset$, $\quad\TS'(\tau^Q)\gets \emptyset$
\For{$i=1\dots n_{tr}$}
\State $b_i \gets \{0 \text{\texttt{ if }}x_{ij}\leq t\text{\texttt{ else }1}\quad\forall t\in\tau_j\ \forall j=1\dots m\}$
\State $\TR'(\tau^Q)\gets \TR'(\tau^Q) \cup \{(b_i, y_i)\}$
\EndFor
\For{$\ell=1\dots n_{ts}$}
\State $b_\ell \gets \{0 \text{\texttt{ if }}x_{\ell j}\leq t\text{\texttt{ else }1}\quad\forall t\in\tau_j\ \forall j=1\dots m\}$
\State $\TS'(\tau^Q)\gets \TS'(\tau^Q) \cup \{(b_\ell, y_\ell)\}$
\EndFor
\State \textbf{return} $\TR'(\tau^Q), \TS'(\tau^Q)$
\end{algorithmic}
\caption{Pseudocode for the FCCA procedure}\label{alg:code}
\end{algorithm}

\section{Experimental Setup}\label{sec:exp_setup}
We tested the procedure summarized by Algorithm \ref{alg:code} on several binary classification datasets with continuous features, whose characteristics are summarized in Table \ref{tab:datasets}. 
As Target black-box model, we can use any black-box model for which it is possible to efficiently solve problem \eqref{eq:CE}. Possible choices are Tree Ensembles (e.g. Random Forest, Gradient Boosting) or Support Vector Machines with a linear kernel. Since the objective is to learn the decision boundaries of $\T$, it is important to choose a model with good performance on the dataset considered. In our analysis, we use Gradient Boosting with 100 estimators of depth 1 and learning rate equal to 0.1: the performance of this algorithm on the datasets computed in $k$-fold crossvalidation is shown in Table \ref{tab:experimental_setting}. The advantage of using such a simple algorithm is that solving problem \eqref{eq:CE_obj}-\eqref{eq:CE_actionability} is extremely fast (order of $10^{-2}$ seconds).

The experiments have been run on an Intel i7-1165G7 2.80GHz CPU with 16GB of available RAM, running Windows 11. The procedure was implemented in Python by using scikit-learn v1.2.1 for training our models; the optimization problem \eqref{eq:CE_obj}-\eqref{eq:CE_actionability} for computing Counterfactual Explanations was solved with Gurobi 10.0.1 \citep{gurobi}.  The value of $\epsilon$ used in constraint \eqref{eq:CE_class_assignment_d} is set to $10^{-4}$. The code of the experiments is available at \url{https://github.com/ceciliasalvatore/sFCCA.git}.  The experimental settings used in our experiments are described in Table \ref{tab:experimental_setting}.

\begin{table}[h]
    \centering
    \begin{tabular}{cccccc}
         Name & $n$ & $m$ & \# training points & $k$ & \# external test points \\\hline
         \textit{boston} & 506 & 14 &506 & 5 & 0\\
         \textit{arrhythmia} & 453& 191& 453 & 5 & 0\\
         \textit{ionosphere} & 351& 32 & 351 & 5 &0\\
         \textit{magic} & 19020 & 10 & 5000 & 5 & 14020\\
         \textit{particle} & 86209& 50&5000 & 5 & 81209\\
         \textit{vehicle} & 98928 & 100 &5000 & 5 &93928\\
    \end{tabular}
    \caption{Summary of the datasets used in the experimental phase. We report the dataset name; the total number of data points $n$; the number of features $m$; the number of data points used as training set; the value of $k$ used for the $k$-fold crossvalidation; the number of data points used as an external test set.} 
    \label{tab:datasets}
\end{table}

As shown in Table \ref{tab:datasets}, we considered both small datasets with less than 5000 data points (\textit{boston}, \textit{arrhythmia} and \textit{ionosphere}) and big datasets with more than 5000 data points (\textit{magic}, \textit{particle}, \textit{vehicle}). For each dataset we compute the performance in a 5-fold crossvalidation. For big datasets we restrict the dataset to a random subset of 5000 data points; the remaining ones are used as an external test set to additionally validate the performance of the system. 

The procedure requires setting a group of parameters:
\begin{itemize}
    \item $p_0$: lower bound on the classification probability for points in the set ${\cal M}$. We recommend using $p_0=0.5$. A higher value can be set if the dataset is very noisy or unbalanced.
    \item $p_1$: upper bound on the classification probability for points in the set ${\cal M}$. We recommend using $p_1=1.0$. A lower value can be used to reduce the number of points of which computing the Counterfactual Explanation if \textit{a)} the cardinality of ${\cal M}$ is large (more than 1000 points) or \textit{b)} the Target model $\T$ is complex (e.g. it is an ensemble of trees with high depth) so that the time for computing each Counterfactual Explanation can be high. 
    \item $\lambda_0$, $\lambda_1$, $\lambda_2$: hyperparameters for the Counterfactual Explanation problem \eqref{eq:CE_obj}-\eqref{eq:CE_actionability}. We recommend setting $\lambda_2=0$ in order to significantly reduce the computational time for the Counterfactual Explanation problem. $\lambda_0$ and $\lambda_1$ must be set to trade-off between sparsity and proximity; we can set $\lambda_1=1$ without loss of generality so that we only need to tune $\lambda_0$. By setting $\lambda_0=0.1$ and $\lambda_1=1$, we mean that changing one additional feature has the same weight as changing the absolute value of one feature of 0.1, which is the 10\% of the scale range (recall that we assume features to be scaled between 0 and 1).
    \item $Q$: the value of $Q$ represents the granularity for the thresholds selected. The standard value is $Q=0$, meaning that we consider all the thresholds extracted from the Counterfactual Analysis. Increasing the value of $Q$ makes the dataset more sparse, but, at the same time, it can degrade the input-output relationship. The trade-off between compression rate and inconsistency rate must be carefully considered when $Q$ is increased.
\end{itemize}

\begin{table}[ht]
    \centering
    \begin{tabular}{cccccccc}
        Name & $p_0$ & $p_1$ & $\lambda_0$ & $\lambda_1$ & $\lambda_2$ & accuracy ${\T}$ (\%) & time (sec.)\\\hline
        \textit{boston} & 0.5 & 1 & 0.1 & 1 & 0 & 82.81 & 7.73\\
        \textit{arrhythmia} & 0.5 & 1 & 0.1 & 1 & 0 & 79.44 & 21.54\\
        \textit{ionosphere} & 0.5 & 1 & 0.1 & 1 & 0 & 91.46 & 8.15\\
        \textit{magic} &  0.5 & 0.7 & 0.1 & 1 & 0 & 83.13 & 23.81\\
        \textit{particle} & 0.5 & 0.7 & 0.1 & 1 & 0 & 88.94 & 16.80\\
        \textit{vehicle} & 0.5 & 0.7 & 0.1 & 1 & 0 & 85.43 & 25.23\\
    \end{tabular}
    \caption{Experimental setting in the different dataset. The Target model $\T$ used is the Gradient Boosting algorithm with 100 estimators and depth 1. We also report the computational time needed to run the FCCA procedure on a single fold of the dataset.}
    \label{tab:experimental_setting}
\end{table}

\subsection{Performance Evaluation}
In order to evaluate the effectiveness of the FCCA procedure, we look at two points of view:
\begin{itemize}
    \item The performance of a classification model trained on the discretized dataset. In particular, our objective is to train a small Decision Tree; we can train Decision Trees either by using a heuristic algorithm such as CART or by using an optimal approach. In the second case, the discretization procedure allows us to use the GOSDT approach proposed by \cite{pmlr-v119-lin20g}, which requires binary features in input. For both CART and GOSDT, we limit to 3 the maximum depth allowed. For GOSDT, we also set the regularization parameter to $\frac{10}{n_{tr}}$; in our experiments, in fact, we noticed that using this value allows us to produce Decision Trees that are shallow but can properly represent the input-output relationship.
    \item The compression and inconsistency rates of the discretized dataset.
\end{itemize}
Results obtained by the FCCA procedure are compared with the ones obtained by the initial dataset with continuous features and by the dataset discretized following the ``Guessing Thresholds'' procedure proposed in \cite{McTavish_Zhong_Achermann_Karimalis_Chen_Rudin_Seltzer_2022}, that we will refer to as GTRE (Guessing Thresholds via Reference Ensemble). The GTRE procedure is also based on the idea of leveraging a Tree Ensemble model to extract relevant thresholds that discretize the input dataset in a meaningful way; the GTRE procedure uses as Tree Ensemble the Gradient Boosting algorithm. In all the experiments, when we run the two methods we set the same parameters for the Gradient Boosting to have a fair comparison (100 trees of depth 1 and learning rate 0.1).

\subsection{Results}\label{sec:results}
In this section, we analyze the results obtained in our experiments in terms of accuracy, sparsity, compression rate and inconsistency rate.

\subsubsection{Accuracy}\label{sec:accuracy}
\begin{figure}
    \centering
    \begin{subfigure}[b]{0.45\textwidth}
        \centering
        \includegraphics[width=\textwidth]{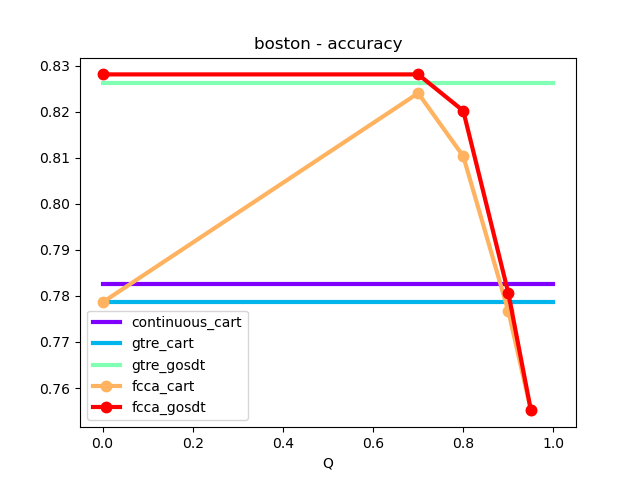}
        \caption{\textit{boston}}
        \label{fig:acc_boston}
    \end{subfigure}
    \hfill
    \begin{subfigure}[b]{0.45\textwidth}
        \centering
        \includegraphics[width=\textwidth]{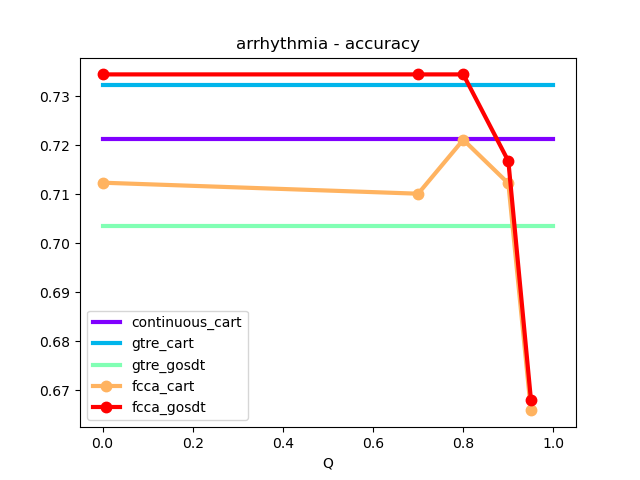}
        \caption{\textit{arrhythmia}}
        \label{fig:acc_arrhythmia}
    \end{subfigure}
    \hfill
        \begin{subfigure}[b]{0.45\textwidth}
        \centering
        \includegraphics[width=\textwidth]{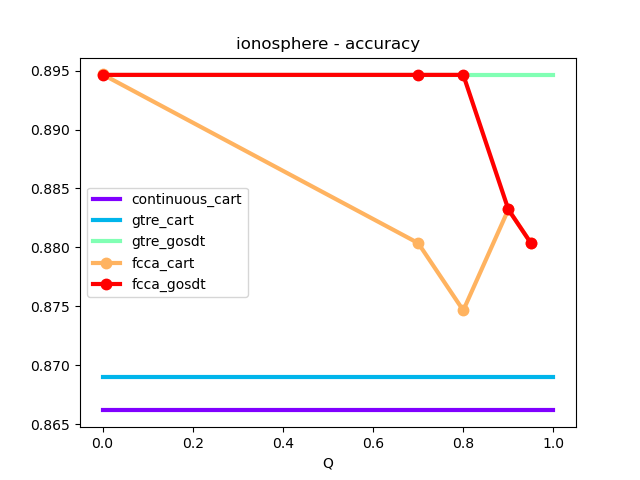}
        \caption{\textit{ionosphere}}
        \label{fig:acc_ionosphere}
    \end{subfigure}
    \hfill
        \begin{subfigure}[b]{0.45\textwidth}
        \centering
        \includegraphics[width=\textwidth]{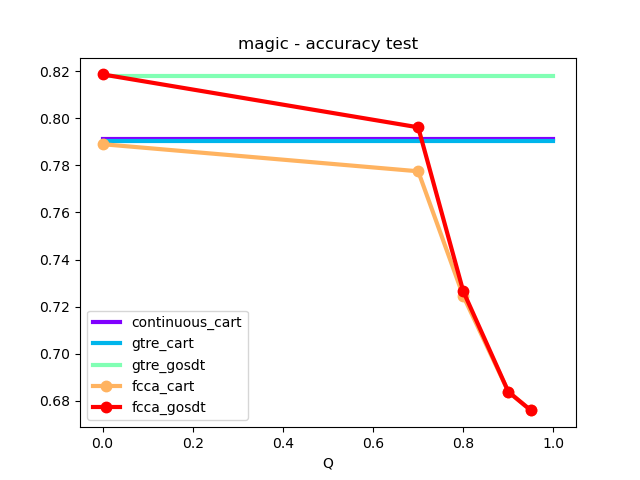}
        \caption{\textit{magic}}
        \label{fig:acc_magic}
    \end{subfigure}
    \hfill
        \begin{subfigure}[b]{0.45\textwidth}
        \centering
        \includegraphics[width=\textwidth]{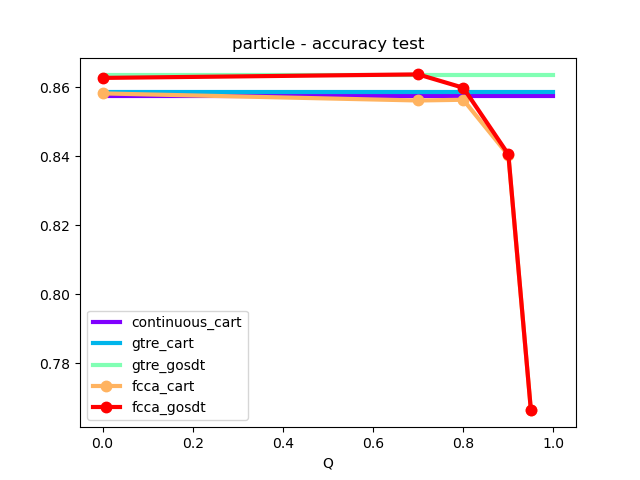}
        \caption{\textit{particle}}
        \label{fig:acc_particle}
    \end{subfigure}
    \hfill
        \begin{subfigure}[b]{0.45\textwidth}
        \centering
        \includegraphics[width=\textwidth]{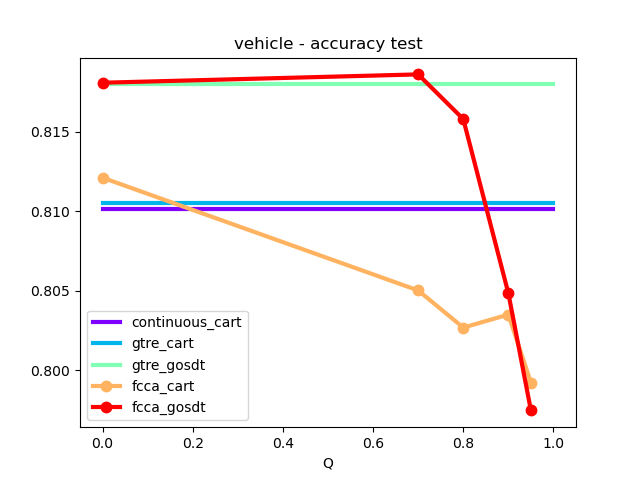}
        \caption{\textit{vehicle}}
        \label{fig:acc_vehicle}
    \end{subfigure}
    \hfill
    \caption{Accuracy results on the benchmark datasets. We compare the performance of CART and GOSDT trained on the initial dataset with continuous features, the dataset discretized with the GTRE procedure, and the dataset discretized with the FCCA procedure. It is not possible to apply GOSDT directly to the dataset with continuous features. As reported in Table \ref{tab:datasets}, for datasets with few observations (\textit{boston}, \textit{arrhythmia} and \textit{ionosphere}) the accuracy is computed in a $k$-fold crossvalidation, while for datasets with many observations (\textit{magic}, \textit{particle} and \textit{vehicle}) the accuracy is computed as the average result of the $k$ classifiers trained in $k$-fold crossvalidation on the external test set.}
    \label{fig:accuracy}
\end{figure}

\begin{figure}
    \centering
    \begin{subfigure}[b]{0.45\textwidth}
        \centering
        \includegraphics[width=\textwidth]{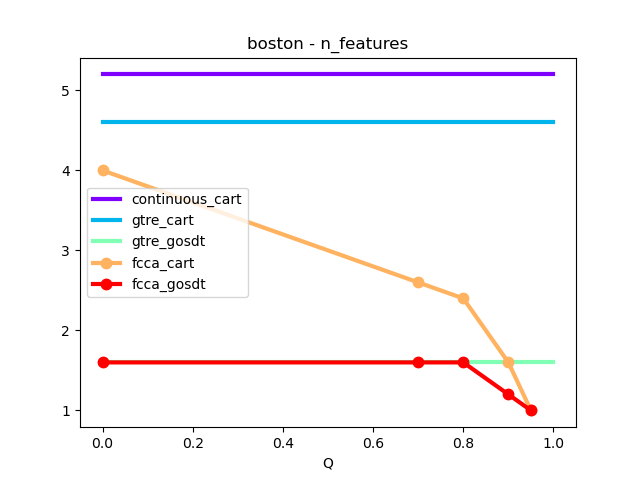}
        \caption{\textit{boston}}
        \label{fig:n_features_boston}
    \end{subfigure}
    \hfill
    \begin{subfigure}[b]{0.45\textwidth}
        \centering
        \includegraphics[width=\textwidth]{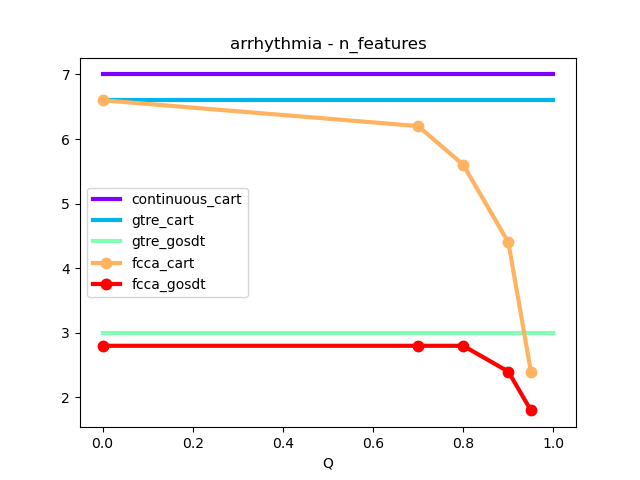}
        \caption{\textit{arrhythmia}}
        \label{fig:n_features_arrhythmia}
    \end{subfigure}
    \hfill
        \begin{subfigure}[b]{0.45\textwidth}
        \centering
        \includegraphics[width=\textwidth]{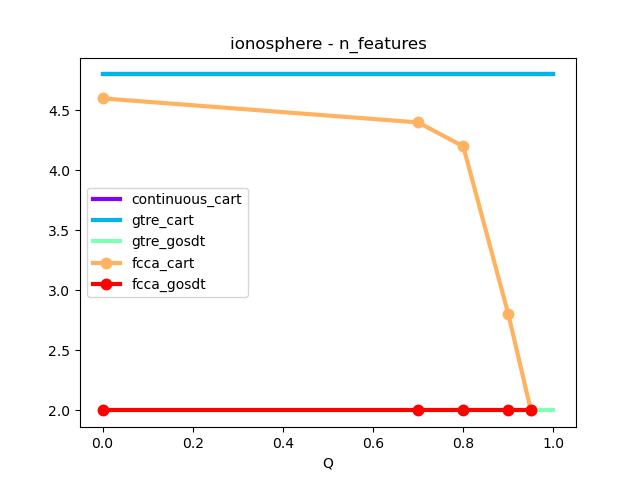}
        \caption{\textit{ionosphere}}
        \label{fig:n_features_ionosphere}
    \end{subfigure}
    \hfill
        \begin{subfigure}[b]{0.45\textwidth}
        \centering
        \includegraphics[width=\textwidth]{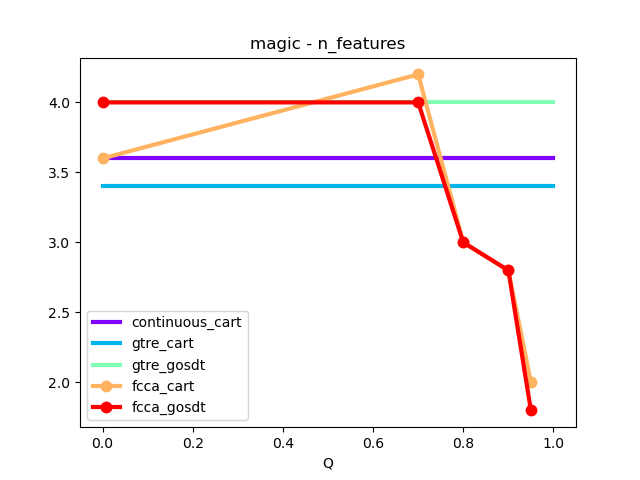}
        \caption{\textit{magic}}
        \label{fig:n_features_magic}
    \end{subfigure}
    \hfill
        \begin{subfigure}[b]{0.45\textwidth}
        \centering
        \includegraphics[width=\textwidth]{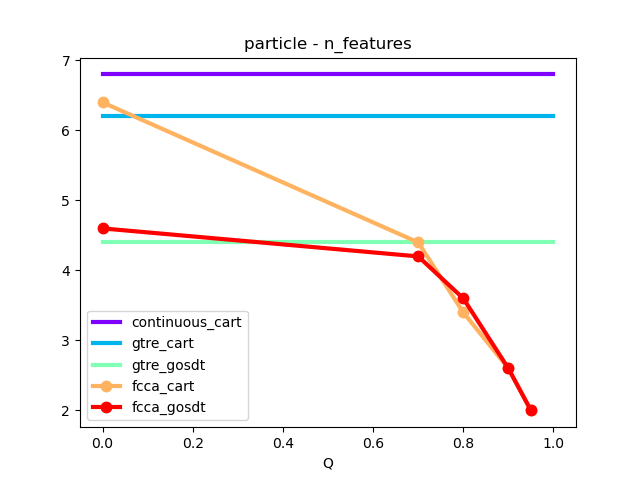}
        \caption{\textit{particle}}
        \label{fig:n_features_particle}
    \end{subfigure}
    \hfill
        \begin{subfigure}[b]{0.45\textwidth}
        \centering
        \includegraphics[width=\textwidth]{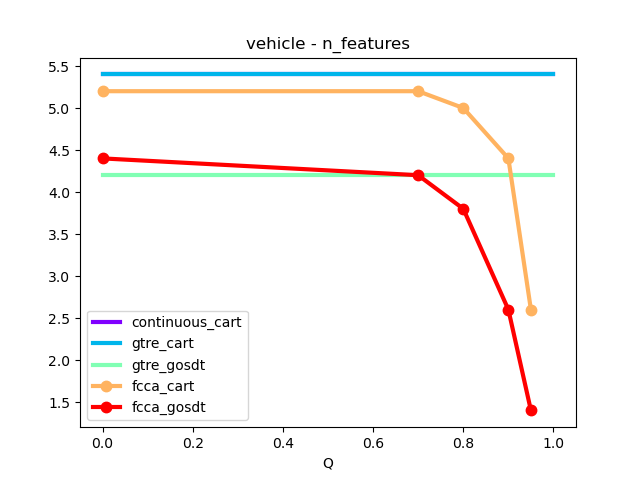}
        \caption{\textit{vehicle}}
        \label{fig:n_features_vehicle}
    \end{subfigure}
    \hfill
    \caption{Number of features used on the benchmark datasets. We compare the performance of CART and GOSDT trained on the initial dataset with continuous features, the dataset discretized with the GTRE procedure, and the dataset discretized with the FCCA procedure.}
    \label{fig:n_features}
\end{figure}

In Figure \ref{fig:accuracy} we present the results in terms of accuracy on the benchmark datasets obtained by training both CART and GOSDT on the initial dataset with continuous features, the dataset discretized with the GTRE procedure, and the dataset discretized with the FCCA procedure at different levels of quantile $Q$. Please note that it is not possible to apply the GOSDT approach directly to the dataset with continuous features: in this case, it would, in fact, be necessary to introduce a threshold in the middle of any two consecutive data points for each feature; the resulting dataset is too big, and it is thus not affordable to compute GOSDT. In all datasets, we can notice that using a discretization technique is beneficial because it allows us to compute the optimal tree with the GOSDT algorithm; using this algorithm results in a more robust, accurate and sparse (Figure \ref{fig:n_features}) decision tree with respect to the heuristic CART. Combining GOSDT with the two discretization techniques GTRE and FCCA for $Q=0$ achieves similar performance on all datasets; increasing the value of $Q$ to 0.7 preserves a high accuracy level on almost all datasets, except for \textit{magic}.

\subsubsection{Compression and Inconsistency Rate}
\begin{figure}
    \centering
    \begin{subfigure}[b]{0.45\textwidth}
        \centering
        \includegraphics[width=\textwidth]{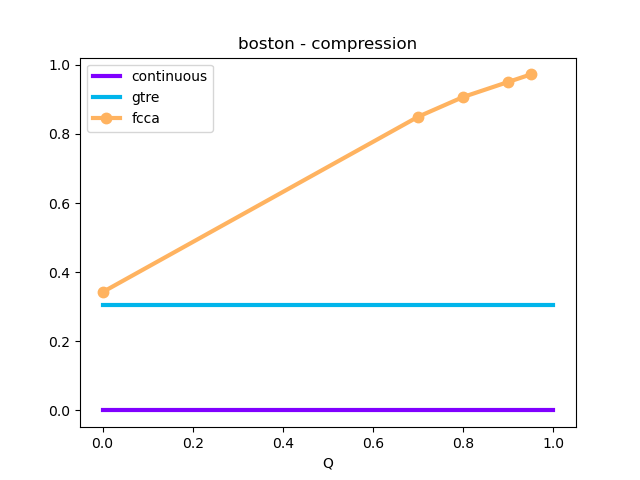}
        \caption{\textit{boston}}
        \label{fig:compression_boston}
    \end{subfigure}
    \hfill
    \begin{subfigure}[b]{0.45\textwidth}
        \centering
        \includegraphics[width=\textwidth]{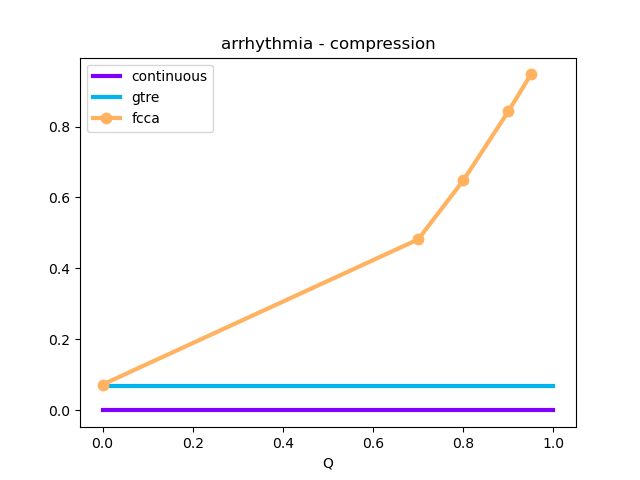}
        \caption{\textit{arrhythmia}}
        \label{fig:compression_arrhythmia}
    \end{subfigure}
    \hfill
        \begin{subfigure}[b]{0.45\textwidth}
        \centering
        \includegraphics[width=\textwidth]{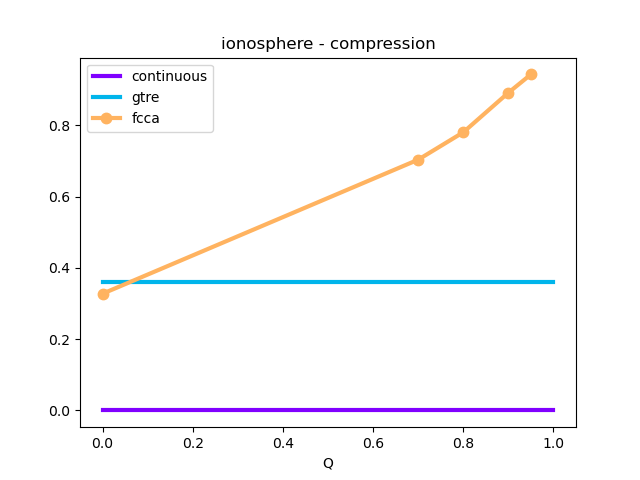}
        \caption{\textit{ionosphere}}
        \label{fig:compression_ionosphere}
    \end{subfigure}
    \hfill
        \begin{subfigure}[b]{0.45\textwidth}
        \centering
        \includegraphics[width=\textwidth]{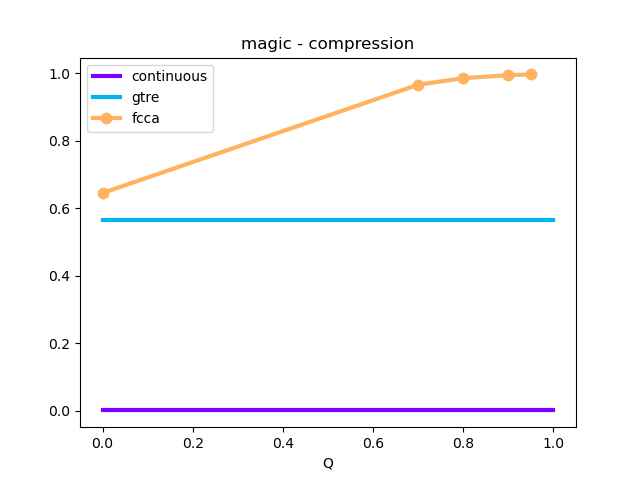}
        \caption{\textit{magic}}
        \label{fig:compression_magic}
    \end{subfigure}
    \hfill
        \begin{subfigure}[b]{0.45\textwidth}
        \centering
        \includegraphics[width=\textwidth]{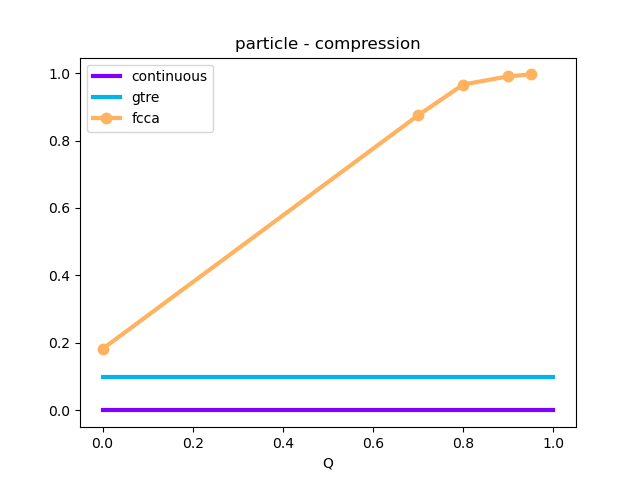}
        \caption{\textit{particle}}
        \label{fig:compression_particle}
    \end{subfigure}
    \hfill
        \begin{subfigure}[b]{0.45\textwidth}
        \centering
        \includegraphics[width=\textwidth]{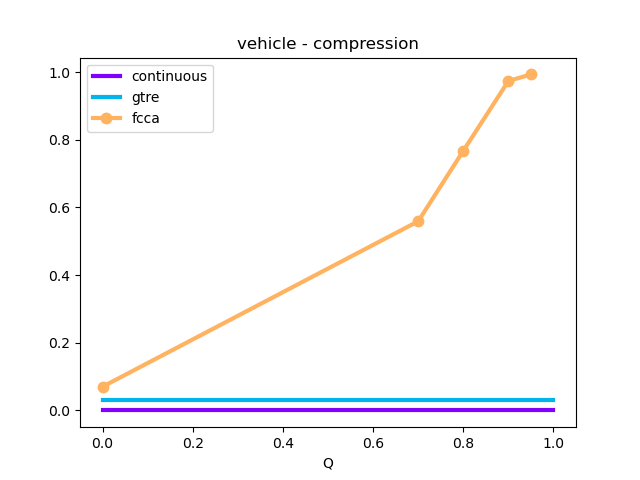}
        \caption{\textit{vehicle}}
        \label{fig:compression_vehicle}
    \end{subfigure}
    \hfill
    \caption{Compression rate on the benchmark datasets. We compare the performance of CART and GOSDT trained on the initial dataset with continuous features, the dataset discretized with the GTRE procedure, and the dataset discretized with the FCCA procedure.}
    \label{fig:compression}
\end{figure}

\begin{figure}
    \centering
    \begin{subfigure}[b]{0.45\textwidth}
        \centering
        \includegraphics[width=\textwidth]{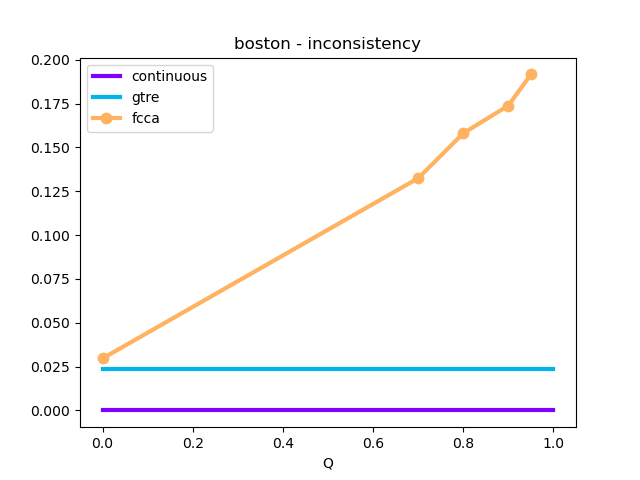}
        \caption{\textit{boston}}
        \label{fig:inconsistency_boston}
    \end{subfigure}
    \hfill
    \begin{subfigure}[b]{0.45\textwidth}
        \centering
        \includegraphics[width=\textwidth]{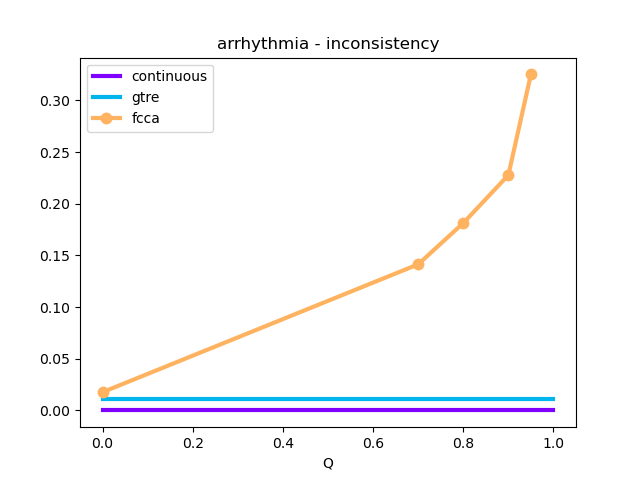}
        \caption{\textit{arrhythmia}}
        \label{fig:inconsistency_arrhythmia}
    \end{subfigure}
    \hfill
        \begin{subfigure}[b]{0.45\textwidth}
        \centering
        \includegraphics[width=\textwidth]{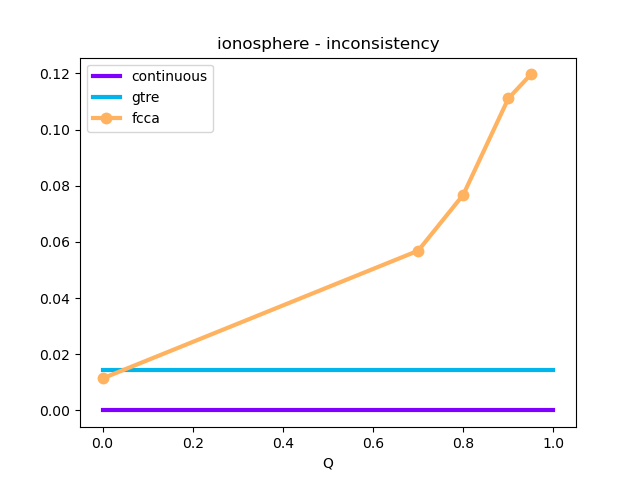}
        \caption{\textit{ionosphere}}
        \label{fig:inconsistency_ionosphere}
    \end{subfigure}
    \hfill
        \begin{subfigure}[b]{0.45\textwidth}
        \centering
        \includegraphics[width=\textwidth]{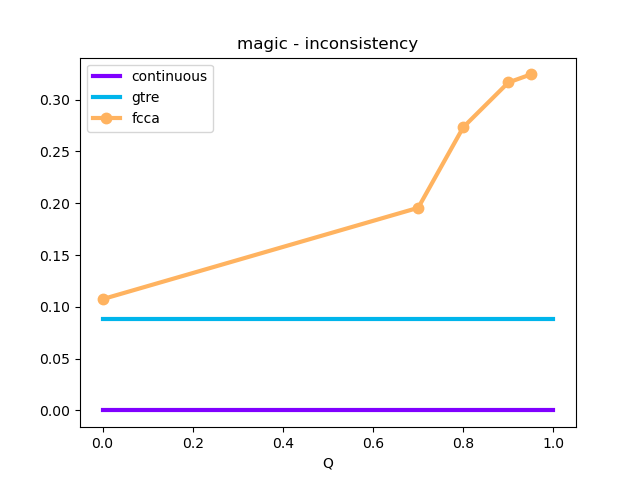}
        \caption{\textit{magic}}
        \label{fig:inconsistency_magic}
    \end{subfigure}
    \hfill
        \begin{subfigure}[b]{0.45\textwidth}
        \centering
        \includegraphics[width=\textwidth]{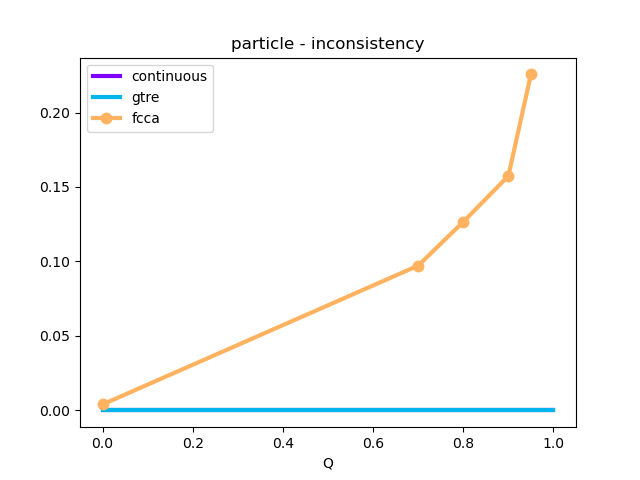}
        \caption{\textit{particle}}
        \label{fig:inconsistency_particle}
    \end{subfigure}
    \hfill
    \begin{subfigure}[b]{0.45\textwidth}
        \centering
        \includegraphics[width=\textwidth]{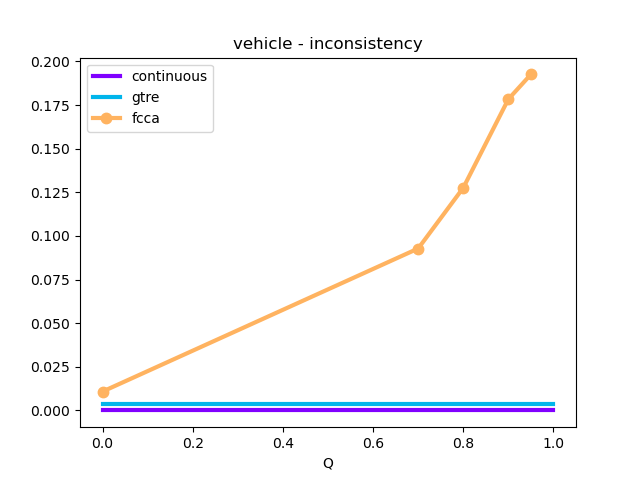}
        \caption{\textit{vehicle}}
        \label{fig:inconsistency_vehicle}
    \end{subfigure}
    \hfill
    \caption{Inconsistency rate on the benchmark datasets. We compare the performance of CART and GOSDT trained on the initial dataset with continuous features, the dataset discretized with the GTRE procedure, and the dataset discretized with the FCCA procedure.}
    \label{fig:inconsistency}
\end{figure}

In Figures \ref{fig:compression} and \ref{fig:inconsistency}, we plot respectively the compression rate $\eta$ and the inconsistency rate $\delta$ of the initial dataset with continuous features, the dataset discretized with the GTRE procedure, and the dataset discretized with the FCCA procedure at different levels of quantile $Q$. On all datasets, we can notice that the initial dataset with continuous features has zero compression and zero inconsistency, while both GTRE and FCCA have a significant compression rate that, as a downside, leads to a non-zero inconsistency. For $Q=0$, the compression and inconsistency rates of FCCA are very similar to the ones obtained by GTRE. In the FCCA procedure, compression and inconsistency rates are proportional to the quantile value $Q$. In fact, when $Q$ tends to 1, also the compression rate $\eta$ tends to 1. A high compression rate is positive because implies that we are able to summarize the data using information at a low granularity. Therefore it is easier to build a small and interpretable decision tree for learning the input-output relationship of this representation of the dataset. As a downside, however, also the inconsistency rate increases when $Q$ tends to $1$. The trade-off between compression rate and inconsistency rate can help us in deciding whether it is possible to increase the level of $Q$ in order to obtain a more sparse dataset without affecting accuracy. In fact, in all datasets except for \textit{magic} we can observe that increasing $Q$ to 0.7 leads to an acceptable level of inconsistency when compared to the accuracy level that is generally reached on that dataset. On the contrary, on \textit{magic} the inconsistency rate is quite high already for $Q=0$, and it arrives at almost 20\% for $Q=0.7$: for this reason, for $Q=0.7$ we record a decrease in the accuracy (Figure \ref{fig:acc_magic}).

\subsubsection{Sparsity}
\begin{figure}
    \centering
    \begin{subfigure}[b]{0.45\textwidth}
        \centering
        \includegraphics[width=\textwidth]{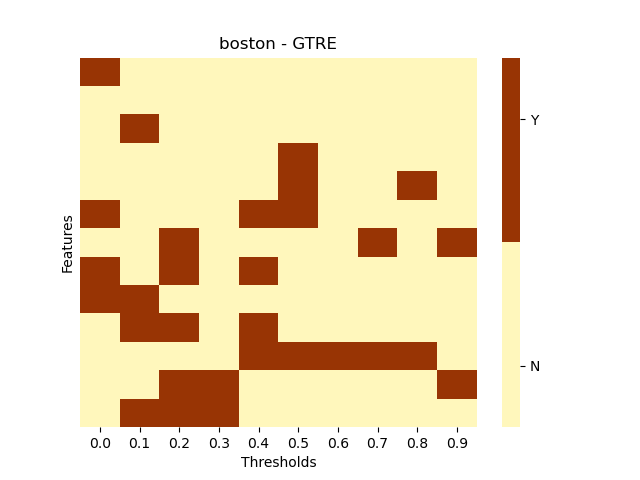}
    \end{subfigure}
    \hfill
    \begin{subfigure}[b]{0.45\textwidth}
        \centering
        \includegraphics[width=\textwidth]{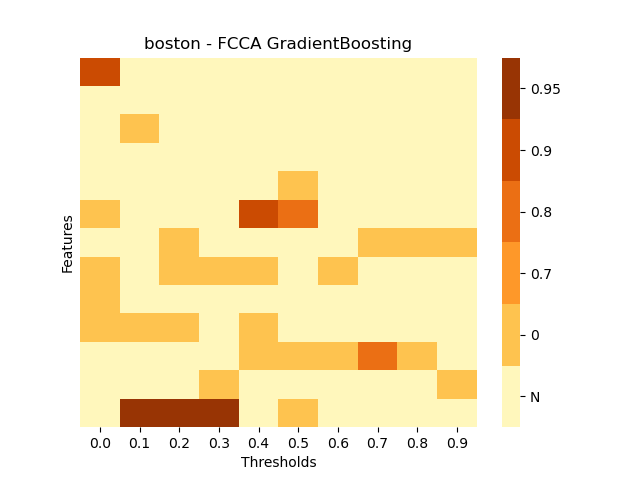}
    \end{subfigure}
    \hfill
    \caption{Heatmap of the thresholds extracted by the GTRE procedure and the FCCA procedure on \textit{boston}. For the GTRE procedure, the heatmap only represents whether, in a given interval, there are some thresholds (Y) or not (N). For the FCCA procedure, and in case there are thresholds in the interval, the heatmap also represents the quantile. Please note that any threshold selected for a given quantile Q will also be selected for lower values of the quantile.}
    \label{fig:thresholds_boston}
\end{figure}

\begin{figure}
    \centering
    \begin{subfigure}[b]{0.45\textwidth}
        \centering
        \includegraphics[width=\textwidth]{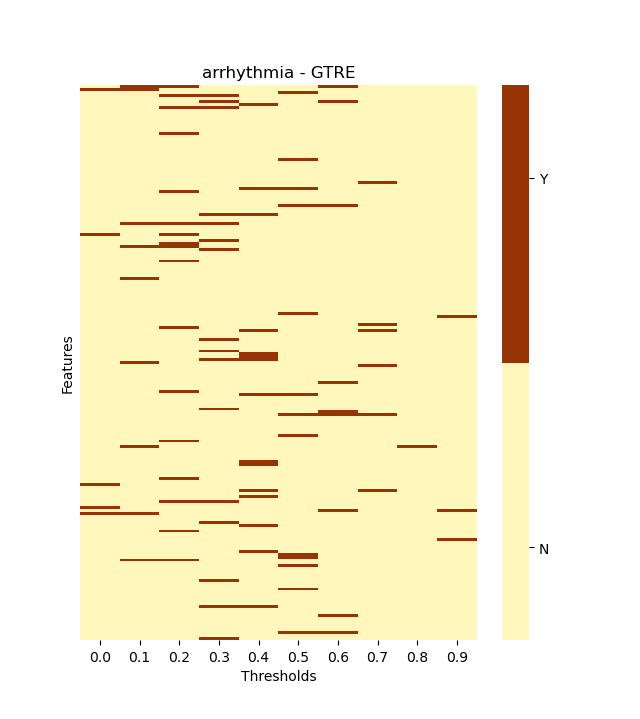}
    \end{subfigure}
    \hfill
    \begin{subfigure}[b]{0.45\textwidth}
        \centering
        \includegraphics[width=\textwidth]{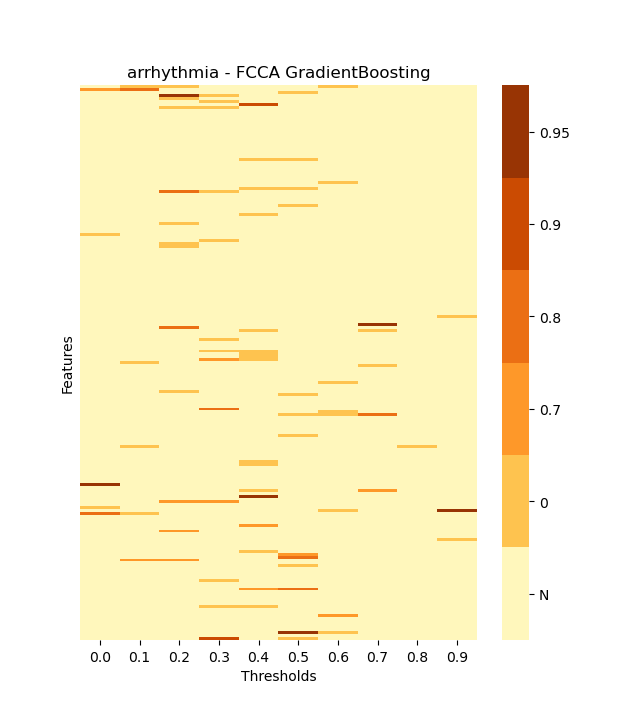}
    \end{subfigure}
    \hfill
    \caption{Heatmap of the thresholds extracted by the GTRE procedure and the FCCA procedure on \textit{arrhythmia}.}
    \label{fig:thresholds_arrhythmia}
\end{figure}

\begin{figure}
    \centering
    \begin{subfigure}[b]{0.45\textwidth}
        \centering
        \includegraphics[width=\textwidth]{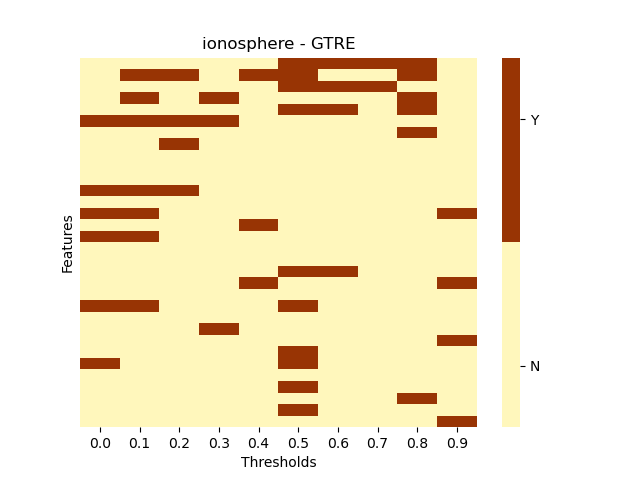}
    \end{subfigure}
    \hfill
    \begin{subfigure}[b]{0.45\textwidth}
        \centering
        \includegraphics[width=\textwidth]{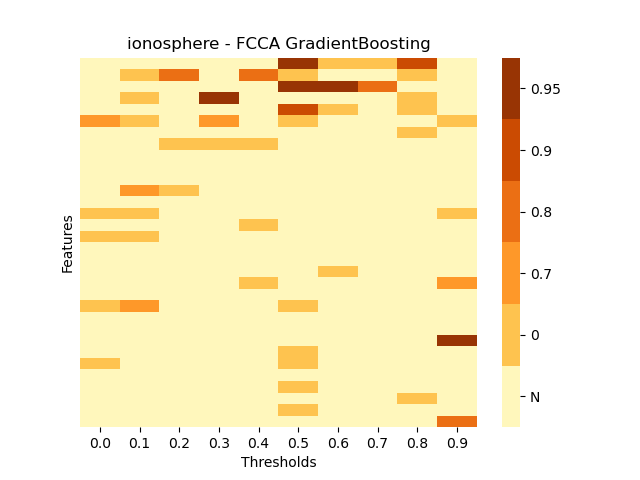}
    \end{subfigure}
    \hfill
    \caption{Heatmap of the thresholds extracted by the GTRE procedure and the FCCA procedure on \textit{ionosphere}.}
    \label{fig:thresholds_ionosphere}
\end{figure}

\begin{figure}
    \centering
    \begin{subfigure}[b]{0.45\textwidth}
        \centering
        \includegraphics[width=\textwidth]{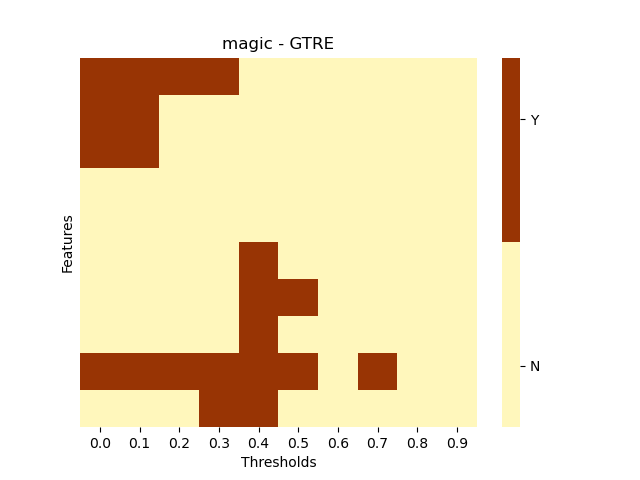}
    \end{subfigure}
    \hfill
    \begin{subfigure}[b]{0.45\textwidth}
        \centering
        \includegraphics[width=\textwidth]{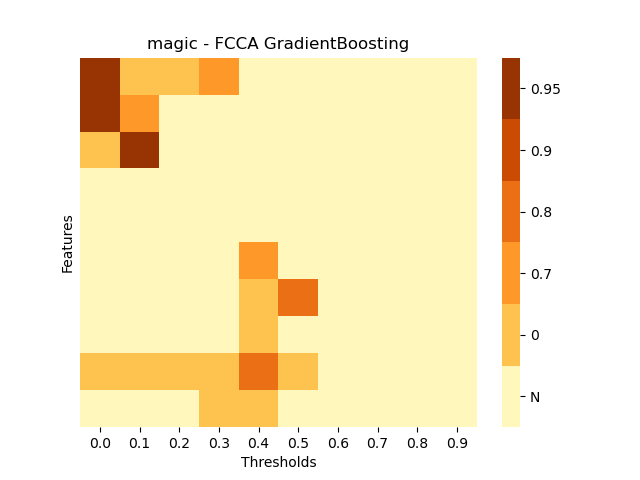}
    \end{subfigure}
    \hfill
    \caption{Heatmap of the thresholds extracted by the GTRE procedure and the FCCA procedure on \textit{magic}.}
    \label{fig:thresholds_magic}
\end{figure}

\begin{figure}
    \centering
    \begin{subfigure}[b]{0.45\textwidth}
        \centering
        \includegraphics[width=\textwidth]{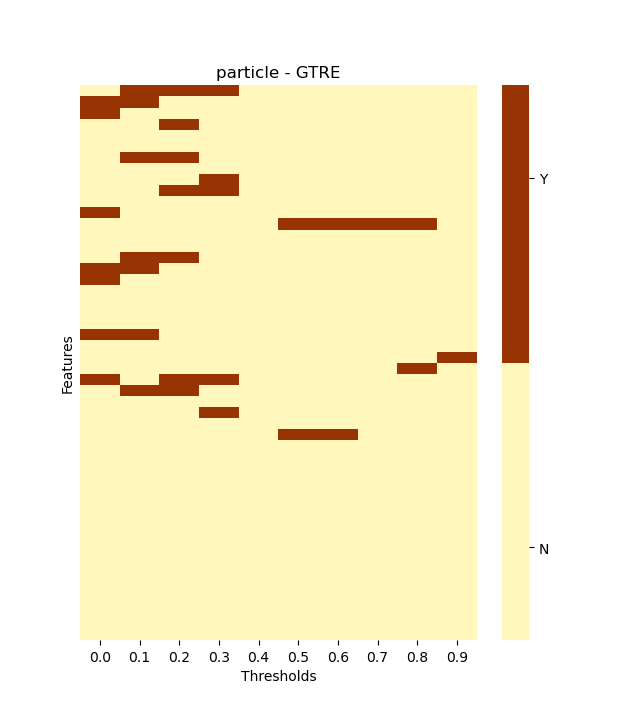}
    \end{subfigure}
    \hfill
    \begin{subfigure}[b]{0.45\textwidth}
        \centering
        \includegraphics[width=\textwidth]{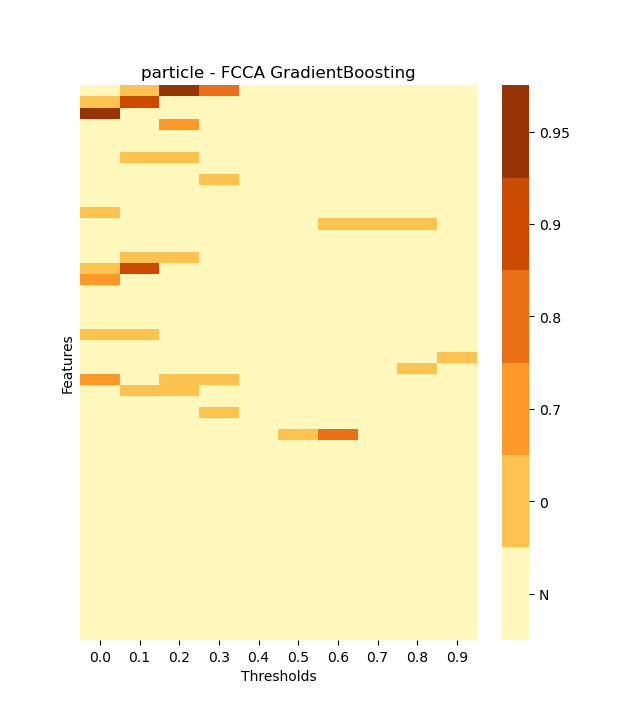}
    \end{subfigure}
    \hfill
    \caption{Heatmap of the thresholds extracted by the GTRE procedure and the FCCA procedure on \textit{particle}.}
    \label{fig:thresholds_particle}
\end{figure}

\begin{figure}
    \centering
    \begin{subfigure}[b]{0.45\textwidth}
        \centering
        \includegraphics[width=\textwidth]{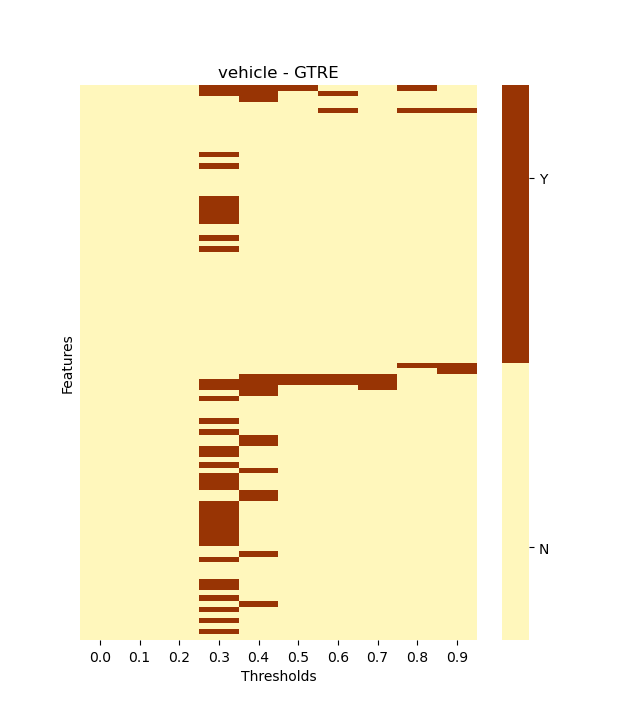}
    \end{subfigure}
    \hfill
    \begin{subfigure}[b]{0.45\textwidth}
        \centering
        \includegraphics[width=\textwidth]{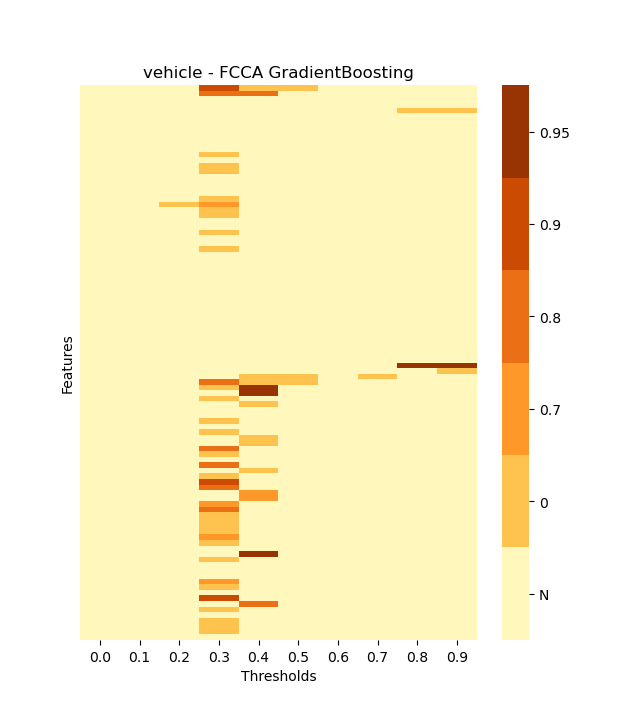}
    \end{subfigure}
    \hfill
    \caption{Heatmap of the thresholds extracted by the GTRE procedure and the FCCA procedure on \textit{vehicle}.}
    \label{fig:thresholds_vehicle}
\end{figure}

As already stated, increasing the value of $Q$ in the FCCA procedure leads to selecting the thresholds that, according to the Counterfactual Analysis, are more relevant in describing the input-output relationship. The discretized dataset derived is thus smaller because it is described through a smaller set of binary features. If $Q$ is too high, we can lose the ability to represent the input-output relationship, ending up building low-performance surrogate classification models. In our previous analysis, we observed that $Q=0.7$ is a good value in almost all of the benchmark datasets considered. In Figures \ref{fig:thresholds_boston}-\ref{fig:thresholds_vehicle}, we analyze the thresholds extracted by the FCCA procedure at different levels of $Q$ and compare them to the thresholds extracted by the GTRE procedure. In these figures, we represent the thresholds extracted by each discretization procedure through a heatmap. For the FCCA procedure, the heatmap also represents the quantile. Please note that thresholds that appear at a certain quantile also appear at lower levels of quantiles. We can notice that the heatmaps obtained through the GTRE procedure and the ones obtained through the FCCA with $Q=0$ are very similar, meaning that the two approaches extract very similar sets of thresholds. When $Q$ grows, instead, the set of thresholds extracted by the FCCA procedure is much smaller.

\subsection{Other Target algorithms}
In this section, we briefly analyze what happens by changing the algorithm implemented by the Target model $\T$. As alternative algorithms to Gradient Boosting, we consider Random Forest or Linear Support Vector Machines. For Random Forest, it is possible to compute Counterfactual Explanations analogously to what is done for Gradient Boosting by solving problem \eqref{eq:CE_obj}-\eqref{eq:CE_actionability}; for Support Vector Machines, instead, Equation \eqref{eq:CE} leads to the following formulation:
\begin{align}
     \min_{x^{CE}}\ & C(x^0, x^{CE}) := \lambda_0\|x^0-x^{CE}\|_0 + \lambda_1\|x^0-x^{CE}\|_1 + \lambda_2\|x^0-x^{CE}\|^2_2\label{eq:SVM_CE_obj}\\  
     &y^{CE}(w^Tx^{CE}+b)\geq 1\label{eq:SVM_CE_class_assignment}\\
     & x^{CE}\in{\cal X}^0.\label{eq:SVM_CE_actionability}
\end{align}

For the sake of brevity, we only show this comparison on one benchmark dataset, \textit{boston}. The hyperparameters for both Random Forest and linear SVMs are computed in crossvalidation: we thus set the maximum depth of the Random Forest to 4 and the value of $C$ for SVM to 1. The other parameters used in this experiment are analogous to the one used in Section \ref{sec:results}.

\begin{figure}
    \centering
    \begin{subfigure}[b]{0.45\textwidth}
        \centering
        \includegraphics[width=\textwidth]{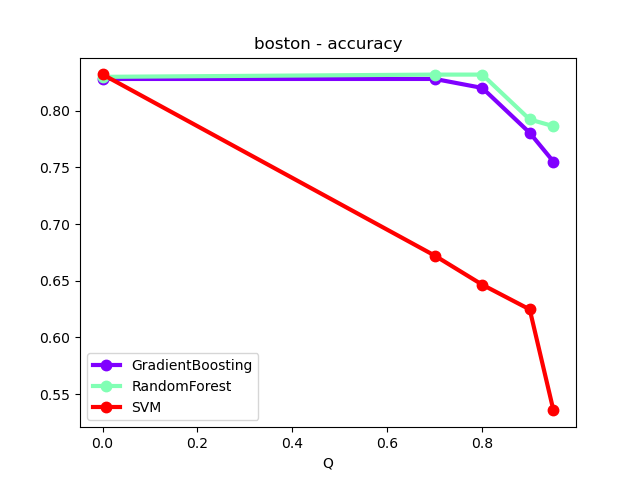}
    \end{subfigure}
    \hfill
    \begin{subfigure}[b]{0.45\textwidth}
        \centering
        \includegraphics[width=\textwidth]{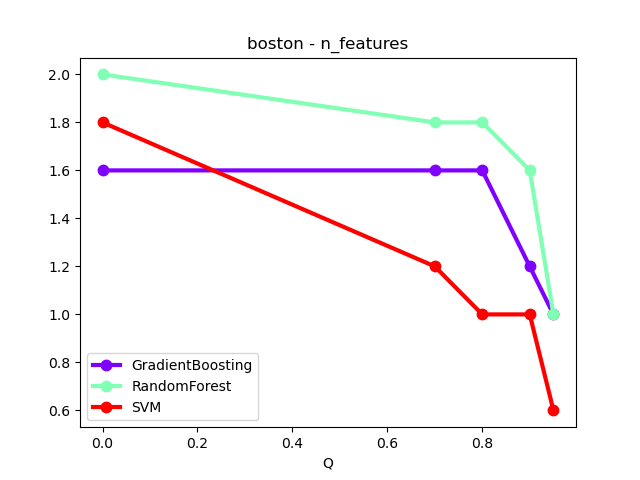}
    \end{subfigure}
    \hfill
    \begin{subfigure}[b]{0.45\textwidth}
        \centering
        \includegraphics[width=\textwidth]{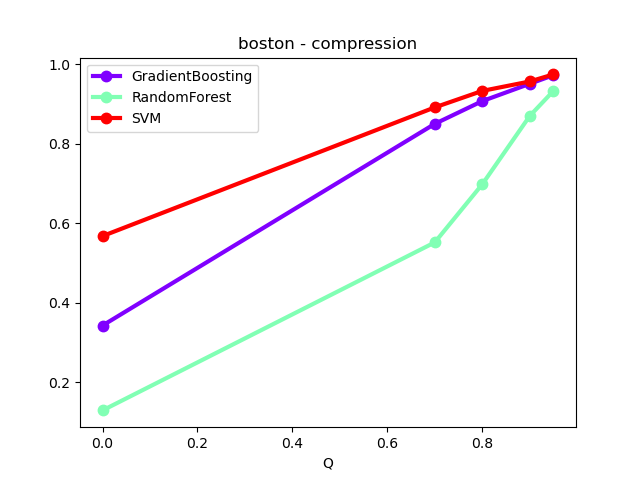}
    \end{subfigure}
    \hfill
    \begin{subfigure}[b]{0.45\textwidth}
        \centering
        \includegraphics[width=\textwidth]{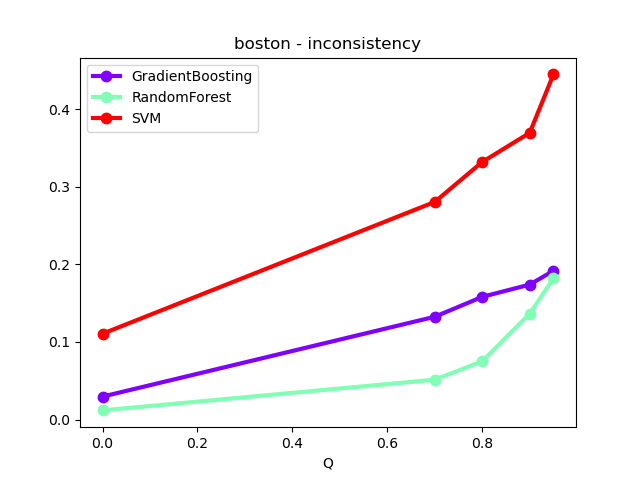}
    \end{subfigure}
    \hfill
    \caption{Comparison on the use of different Target algorithms in the FCCA procedure in terms of accuracy, sparsity, compression and inconsistency on \textit{boston}.}
    \label{fig:FCCA_others}
\end{figure}

\begin{figure}
    \centering
    \begin{subfigure}[b]{0.45\textwidth}
        \centering
        \includegraphics[width=\textwidth]{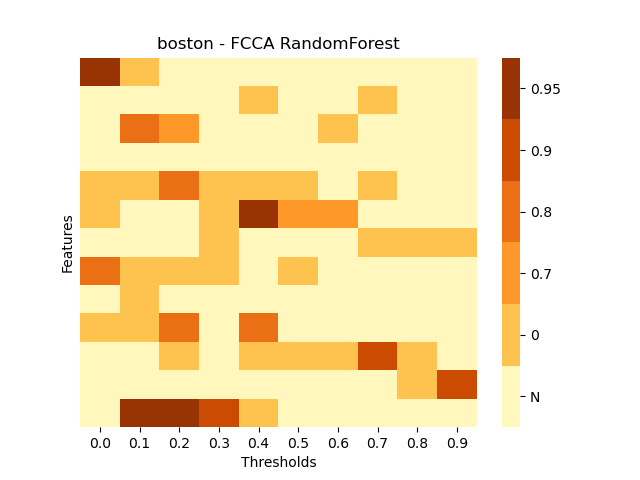}
    \end{subfigure}
    \hfill
    \begin{subfigure}[b]{0.45\textwidth}
        \centering
        \includegraphics[width=\textwidth]{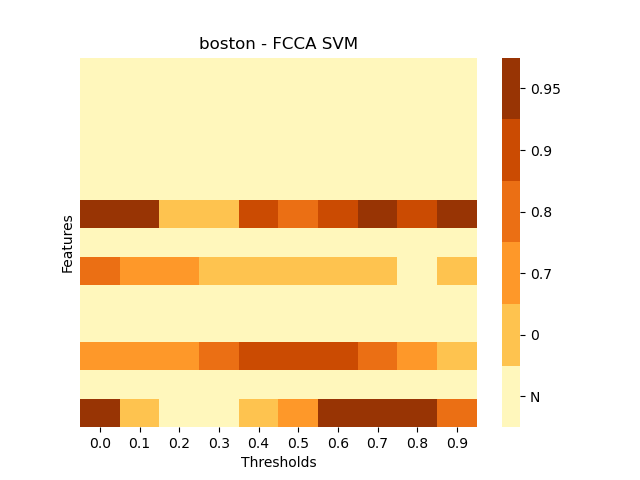}
    \end{subfigure}
    \caption{Heatmap of the thresholds extracted by using the Random Forest and Linear SVM algorithms in the FCCA procedure on \textit{boston}.}
    \label{fig:FCCA_others_heatmaps}
\end{figure}

Figure \ref{fig:FCCA_others} presents the comparison of the FCCA method with different Target algorithms in terms of accuracy, sparsity, compression and inconsistency, while Figure \ref{fig:FCCA_others_heatmaps} presents the heatmap of the thresholds extracted by the FCCA method starting from the Random Forest and Linear SVM algorithms. We can notice that when $Q=0$, the three Target models all lead to similar results in terms of accuracy; however, comparing Figures \ref{fig:thresholds_boston} and \ref{fig:FCCA_others_heatmaps}, we can notice that the set of thresholds extracted is quite different. In particular, while the behaviour for Gradient Boosting and Random Forest is quite similar, for the SVM we can notice that the thresholds extracted are distributed over almost the whole interval of values; this is due to the continuous nature of SVMs. For SVMs, Counterfactual Analysis thus results in a feature selection technique.


\section{Conclusions}\label{sec:concl}
In this paper, we use Counterfactual Analysis to derive a supervised discretization of a dataset driven by a black-box model's classification function. Counterfactual Explanations are computed by solving an optimization problem that identifies the most relevant features and the corresponding cutting points for the black-box model. The Counterfactual Explanations provide a set of univariate decision boundaries that allows discretizing the original dataset into a set of binary variables. Having a compact dataset of binary variables makes it affordable to train an interpretable optimal decision tree by the recent approach proposed in \cite{pmlr-v119-lin20g}. 
Our procedure allows us to discretize the dataset with a tunable granularity,  controlled by means of the parameter $Q$ with $0\leq Q \leq 1$. 
A high granularity, corresponding to low values of $Q$, results in a high level of detail in the dataset, where we consider a higher number of features, each represented by a large number of binary variables. A lower granularity, corresponding to high values of $Q$, results in a sparse dataset, where we only select the most relevant features and model each of them with few binary variables.

Tuning the value of $Q$ is needed to trade-off between the performance of a classification model built on this dataset (somehow inversely proportional to $Q$) and its sparsity (that is a measure of interpretability and directly proportional to $Q$).

In the numerical section, we demonstrate the viability of our method on several datasets different in size, both in terms of the number of data points and the number of features.
We compare our approach with the one proposed in \cite{McTavish_Zhong_Achermann_Karimalis_Chen_Rudin_Seltzer_2022}, where they discretize the dataset by extracting thresholds retraining smaller and smaller version of the original black-box model until the accuracy drops. Our method has the advantage to detect the thresholds by solving an optimization problem, allowing us to tune both the sparsity in terms of features, controlling the cost function of the optimization problem, and the sparsity in terms of the number of thresholds, controlling the parameter $Q$.

As a future line of research, we plan to use the optimization problem defined in \cite{carrizosaGCE23} for computing counterfactuals of groups of individuals. We can add fairness and plausibility constraints to this problem to extract ``fair'' decision boundaries, and hence indirectly impose fairness of the optimal decision tree built by means of those boundaries.

Finally, we plan to formulate the problem of finding an optimal supervised discretization by a combinatorial optimization problem and study efficient algorithms for solving it.

\section*{Acknowledgements}
This research has been financed in part by research project EC H2020 MSCA RISE NeEDS (Grant agreement ID: 822214). This support is gratefully acknowledged. Veronica Piccialli acknowledges financial support from PNRR MUR project PE0000013-FAIR.

\bibliographystyle{apalike}
\bibliography{references.bib}

\end{document}